\documentclass[10pt,twocolumn,letterpaper]{article}

\usepackage{cvpr}
\usepackage{times}
\usepackage{epsfig}
\usepackage{graphicx}
\usepackage{amsmath,mathtools}
\usepackage{amssymb}
\usepackage[ruled,vlined]{algorithm2e}
\usepackage{multirow}
\usepackage{subcaption}
\usepackage{setspace}
\usepackage{booktabs}
\usepackage{changepage}
\usepackage{wrapfig}
\usepackage{enumitem}
\usepackage[toc,page]{appendix}
\allowdisplaybreaks

\usepackage{array}

\usepackage[pagebackref=true,breaklinks=true,letterpaper=true,colorlinks,bookmarks=false]{hyperref}

\cvprfinalcopy 

\def\cvprPaperID{3123} 

\ifcvprfinal\fi

\begin{document}

\title{Grid-GCN for Fast and Scalable Point Cloud Learning}
\author{\hspace*{-23pt}Qiangeng Xu$^{1}$ \qquad Xudong Sun$^{2}$ \qquad Cho-Ying Wu$^{1}$ \qquad Panqu Wang$^{2}$ \qquad Ulrich Neumann$^1$ \\
    \hspace{-15mm}$^1$University of Southern California \hspace{30mm} $^2$Tusimple, Inc\\
    {\tt\small \hspace{0mm}\{qiangenx,choyingw,uneumann\}@usc.edu}\hspace{5mm}{\tt\small \{xudong.sun,panqu.wang\}@tusimple.ai}\qquad
}
\maketitle
\begin{abstract}
    \vspace{-5pt}
    Due to the sparsity and irregularity of the point cloud data, methods that directly consume points have become popular. Among all point-based models, graph convolutional networks (GCN) lead to notable performance by fully preserving the data granularity and exploiting point interrelation. However, point-based networks spend a significant amount of time on data structuring (e.g., Farthest Point Sampling (FPS) and neighbor points querying), which limit the speed and scalability. In this paper, we present a method, named Grid-GCN, for fast and scalable point cloud learning. Grid-GCN uses a novel data structuring strategy, Coverage-Aware Grid Query (CAGQ). By leveraging the efficiency of grid space, CAGQ improves spatial coverage while reducing the theoretical time complexity. Compared with popular sampling methods such as Farthest Point Sampling (FPS) and Ball Query, CAGQ achieves up to $50\times$ speed-up. With a Grid Context Aggregation (GCA) module, Grid-GCN achieves state-of-the-art performance on major point cloud classification and segmentation benchmarks with significantly faster runtime than previous studies. Remarkably, Grid-GCN achieves the inference speed of \textbf{50fps} on ScanNet using \textbf{81920} points as input. The supplementary \footnote{\href{https://xharlie.github.io/papers/GGCN_supCamReady.pdf}{https://xharlie.github.io/papers/GGCN\textunderscore supCamReady.pdf}} and the code \footnote{\href{https://github.com/xharlie/Grid-GCN}{https://github.com/xharlie/Grid-GCN}} are released.
\end{abstract}
\vspace{-15pt}
\section{Introduction}
    \begin{figure}[!hbt]
        \vspace{-15pt}
        \begin{adjustwidth}{-10pt}{0pt}
            \centering
            \includegraphics[width=1.02\linewidth]{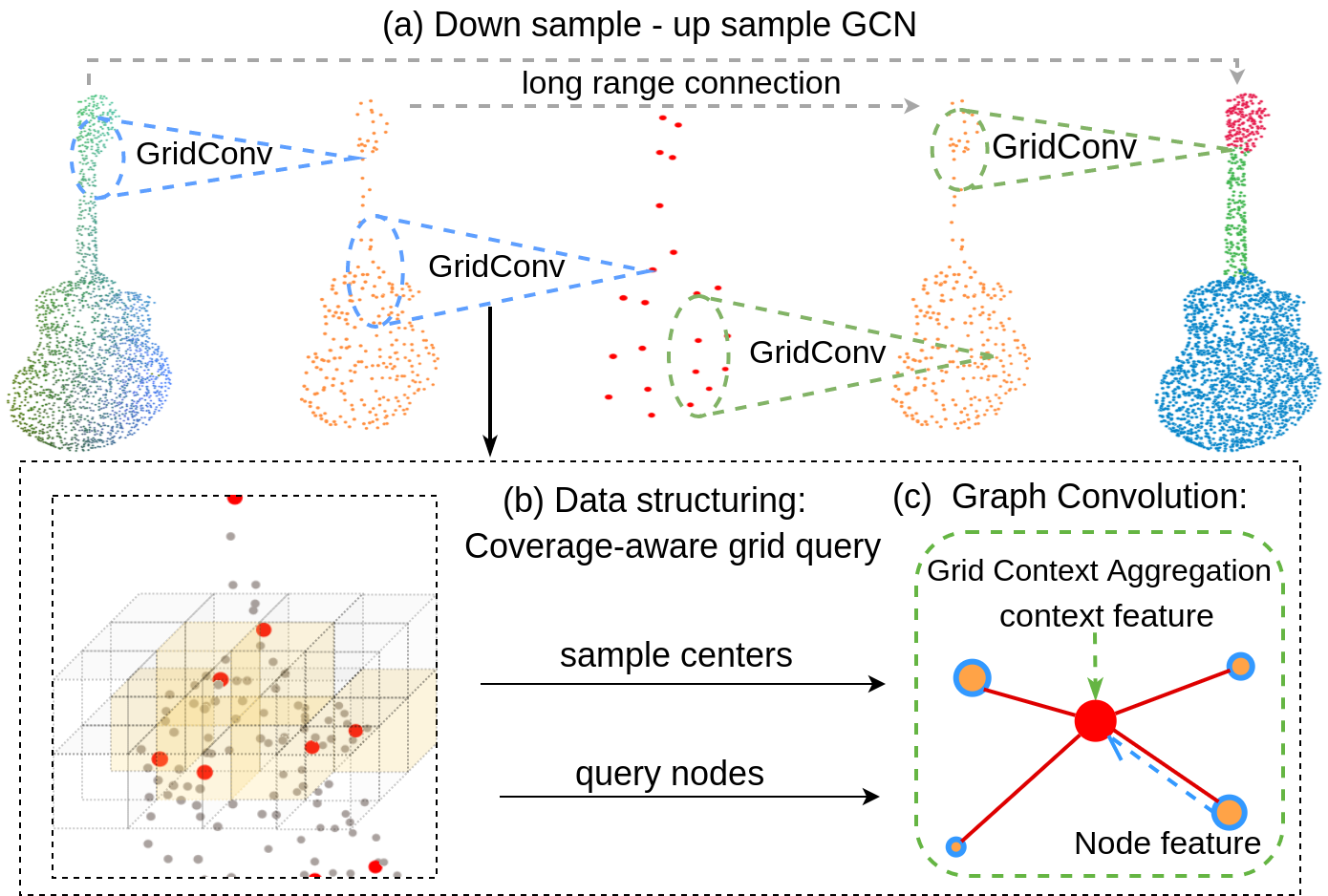}
            \captionsetup{belowskip=-20pt}
            \caption{
                Overview of the Grid-GCN model. (a) Illustration of the network architecture for point cloud segmentation. Our model consists of several GridConv layers, and each can be used in either a downsampling or an upsampling process. A GridConv layer includes two stages: (b) For the data structuring stage, a Coverage-Aware Grid Query (CAGQ) module achieves efficient data structuring and provides point groups for efficient computation. (c) For the convolution stage, a Grid Context Aggregation (GCA) module conducts graph convolution on the point groups by aggregating local context.}
            \label{fig:GGCN}
        \end{adjustwidth}
    \end{figure}
    Point cloud data is popular in applications such as autonomous driving, robotics, and unmanned aerial vehicles. Currently, LiDAR sensors can generate millions of points a second, providing dense real-time representations of the world. Many approaches are used for point cloud data processing. \textit{Volumetric models} are a family of models that transfer point cloud to spatially quantized voxel grids and use a volumetric convolution to perform computation in the grid space \cite{maturana2015voxnet, xu20163d, maturana2015voxnet}. Using grids as data structuring methods, volumetric approaches associate points to locations in grids, and 3D convolutional kernels gather information from neighboring voxels. Although grid data structures are efficient, high voxel resolution is required to preserve the granularity of the data location. Since computation and memory usage grows cubically with the voxel resolution, it is costly to process large point clouds. In addition, since approximately $90\%$ of the voxels are empty for most point clouds\cite{zhou2018voxelnet}, significant computation power may be consumed by processing no information.

    Another family of models for point cloud data processing is \textit{Point-based models}. In contrast to volumetric models, point-based models enable efficient computation but suffer from inefficient data structuring. For example, PointNet \cite{qi2017pointnet} consumes the point cloud directly without quantization and aggregates the information at the last stage of the network, so the accurate data locations are intact but the computation cost grows linearly with the number of points. Later studies \cite{qi2017pointnet++,xu2018spidercnn,wang2019dynamic,wang2018local,NIPS2019_8340} apply a downsampling strategy at each layer to aggregate information into point group centers, therefore extracting fewer representative points layer by layer (Figure \ref{fig:GGCN}(a)). More recently, graph convolutional networks (GCN) \cite{simonovsky2017dynamic,wang2019graph,li2019deepgcns,zhang2019linked} are proposed to  build a local graph for each point group in the network layer, which can be seen as an extension of the PointNet++ architecture \cite{qi2017pointnet++}. However, this architecture incurs high data structuring cost (e.g., FPS and k-NN). Liu et al. \cite{liu2019point} show that the data structuring cost in three popular point-based models \cite{li2018pointcnn,xu2018spidercnn,wang2019dynamic} is up to $88\%$ of the overall computational cost. In this paper, we also examine this issue by showing the trends of data structuring overhead in terms of scalability.

    This paper introduces Grid-GCN, which blends the advantages of volumetric models and point-based models, to achieve efficient data structuring and efficient computation at the same time. As illustrated in Figure \ref{fig:GGCN}, our model consists of several \textit{GridConv} layers to process the point data. Each layer includes two stages: a data structuring stage that samples the representative centers and queries neighboring points; a convolution stage that builds a local graph on each point group and aggregates the information to the center. 
    
    To achieve efficient data structuring, we design a \textbf{Coverage-Aware Grid Query (CAGQ)} module, which 1) accelerates the center sampling and neighbor querying, and 2) provides more complete coverage of the point cloud to the learning process. The data structuring efficiency is achieved through voxelization, and the computational efficiency is obtained through performing computation only on occupied areas. We demonstrate CAGQ's outstanding speed and space coverage in Section \ref{sec:anacagq}. 
    
    To exploit the point relationships, we also describe a novel graph convolution module, named \textbf{Grid Context Aggregation (GCA)}. The module performs \textit{Grid context pooling} to extract context features of the grid neighborhood, which benefits the edge relation computation without adding extra overhead.
    
    We demonstrate the Grid-GCN model on two tasks: point cloud classification and segmentation. Specifically, we perform the classification task on the ModelNet40 and ModelNet10 \cite{wu20153d}, and achieve the state-of-the-art overall accuracy of $93.1\%$ (no voting),  while being on average $5\times$ faster than other models. We also perform the segmentation tasks on ScanNet \cite{dai2017scannet} and S3DIS \cite{2017arXiv170201105A} dataset, and achieve $10\times$ speed-up on average than other models. Notably, our model demonstrates its ability on real-time large-scale point-based learning by processing 81920 points in a scene within 20 ms. (see Section \ref{sec:scala}).
    
\section{Related Work}
    \begin{figure*}[!hbt]
        \begin{adjustwidth}{-12pt}{0pt}
            \centering
            \includegraphics[width=1.03\linewidth]{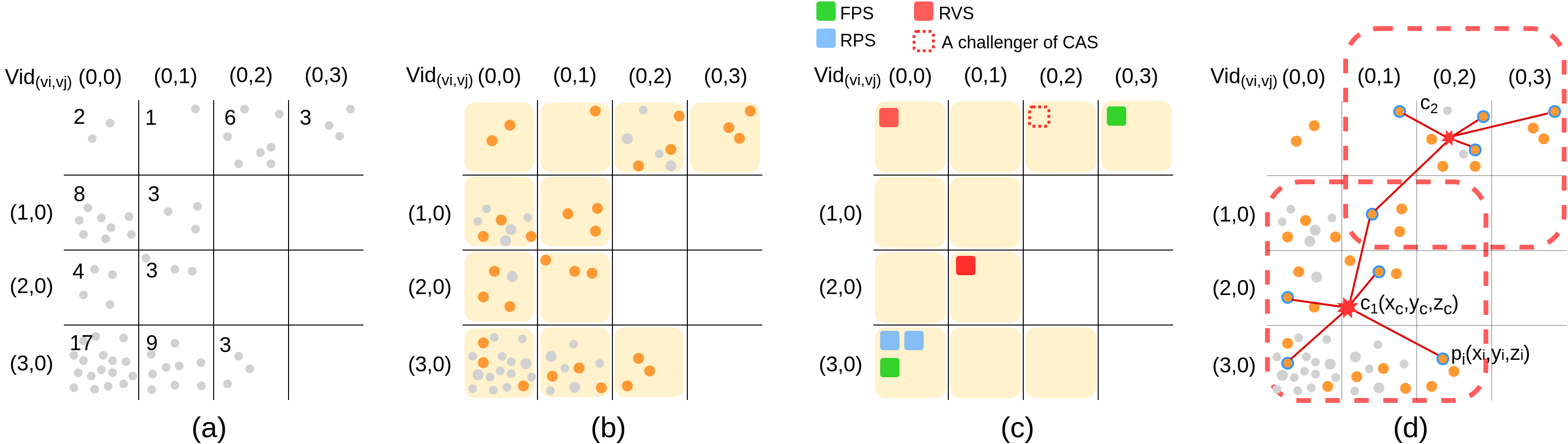}
            \captionsetup{aboveskip=5pt}
            \caption{Illustration of Coverage-Aware Grid Query (CAGQ). Assume we want to sample $M = 2$ point groups and query $K = 5$ node points for each group. \textbf{(a)} The input is $N$ points (grey). The voxel id and number of points is listed for each occupied voxel. \textbf{(b)} We build voxel-point index and store up to $n_v = 3$ points (yellow) in each voxel. \textbf{(c)} Comparison of different sampling methods: FPS and RPS prefer the two centers inside the marked voxels. Our RVS could randomly pick any two occupied voxels (e.g. (2,0) and (0,0)) as center voxels. If our CAS is used, voxel (0,2) will replace (0,0). \textbf{(d)} Context points of center voxel (2,1) are the yellow points in its neighborhood (we use $3\times3$ as an example). CAGQ queries $5$ points (yellow points with blue ring) from these context points, then calculate the locations of the group centers. }
            \label{fig:CAGQ}
        \end{adjustwidth}
        \vspace{-10pt}
    \end{figure*}
    \textbf{Voxel-based methods for 3D learning} To extend the success of convolutional neural network models\cite{he2016deep,huang2017densely} on 2D images, Voxnet and its variants \cite{maturana2015voxnet, wu20153d,wang2015voting,brock2016generative, cciccek20163d} start to transfer point cloud or depth map to occupancy grid and apply volumetric convolution. To address the problem of cubically increased memory usage, OctNet\cite{riegler2017octnet} constructs tree structures for occupied voxels to avoid the computation in the empty space. Although efficient in data structuring, the drawback of the volumetric approach is the low computational efficiency and the loss of data granularity.

    \textbf{Point-based methods for point cloud learning} Point-based models are first proposed by \cite{qi2017pointnet,qi2017pointnet++}, which pursues the permutation invariant by using pooling to aggregate the point features. Approaches such as kernel correlation \cite{atzmon2018point,wu2019pointconv} and extended convolutions \cite{thomas2019kpconv} are proposed to better capture local features. To solve the ordering ambiguity, PointCNN \cite{li2018pointcnn} predicts the local point order, and RSNet \cite{huang2018recurrent} sequentially consumes points from different directions. The computation cost in point-based methods grows linearly with the number of input points. However, the cost of data structuring has become the performance bottleneck on large-scale point clouds.
    
    \textbf{Data structuring strategies for point data} Most point-based methods \cite{qi2017pointnet++,li2018pointcnn,wang2018local,liu2019relation} use FPS \cite{eldar1997farthest} to sample evenly spread group centers. FPS picks the point that maximizes the distance to the selected points. If the number of centers is not very small, the method takes $O(N^2)$ computation. An approximate algorithm \cite{eldar1992irregular} can be $O(NlogN)$. Random Point Sampling (RPS) has the smallest possible overhead, but it's sensitive to density imbalance. Our CAGQ module has the same complexity as RPS, but it performs the sampling and neighbors querying in one shot, which is even faster than RPS with Ball Query or k-NN (see Table \ref{tab:query}). KPConv \cite{thomas2019kpconv} uses a grid sub-sampling to pick points in occupied voxels. Unlike our CAGQ, the strategy cannot query points in the voxel neighbors. CAGQ also has a Coverage-Aware Sampling (CAS) algorithm that optimizes the center selections, which can achieve better coverage than FPS. 
    
    Alternatively, SO-Net \cite{li2018so} builds a self-organizing map. KDNet \cite{klokov2017escape} uses kd-tree to partition the spaces. PATs\cite{yang2019modeling} uses Gumble Subset Sampling to replace FPS. SPG \cite{landrieu2018large} uses a clustering method to group points as super points. All of these methods are either slow in speed or need structure preprocessing. The lattice projection in SPLATNet \cite{su2018splatnet, gu2019hplflownet} preserves more point details than voxel space, but it is slower. Studies such as VoxelNet \cite{zhou2018voxelnet,le2018pointgrid} combines the point-based and volumetric methods by using PointNet \cite{qi2017pointnet} inside each voxel and applying voxel convolution. A concurrent high-speed model PVCNN \cite{liu2019point} uses similar approaches but does not reduce the number of points in each layer progressively. Grid-GCN, yet, can down-sample a large number of points through CAGQ, and aggregate information by considering node relationships in local graphs. 

    \textbf{GCN for point cloud learning} Graph convolutional networks have been widely applied on point cloud learning \cite{wang2019dynamic,landrieu2019point,lan2019modeling}. A local graph is usually built for each point group, and GCN aggregates point data according to relations between points. SpecConv\cite{wang2018local} blends the point features by using a graph Fourier transformation. Other studies model the edge feature between centers and nodes. Among them, \cite{xu2018spidercnn,liu2019relation,lan2019modeling,wang2019dynamic,zhang2019linked} use the geometric relations, while \cite{cciccek20163d,wang2019graph} explore semantic relations between the nodes. Apart from those features, our proposed Grid Context Aggregation module considers coverage and extracts the context features to compute the semantic relation. 
    
\section{Methods}
    \subsection{Method Overview}
    As shown in Figure \ref{fig:GGCN}, Grid-GCN is built on a set of GridConv layers. Each GridConv layer processes the information of $N$ points and maps them to $M$ points. The downsampling GridConv ($N>M$) is repeated several times until a final feature representation is learned. This representation can be directly used for tasks such as classification or further up-sampled by the upsampling GridConv layers ($N < M$) in segmentation tasks. 
    
    GridConv consists of two modules: 
    
    1. A Coverage-Aware Grid Query (CAGQ) module that samples $M$ point groups from $N$ points. Each group includes $K$ node points and a group center. In the upsampling process, CAGQ takes centers directly through long-range connections, and only queries node points for these centers.
    
    2. A Grid Context Aggregation (GCA) module that builds a local graph for each point group and aggregates the information to the group centers. The $M$ group centers are passed as data points for the next layer. 
    
    We list all the notations in the supplementary for clarity.
    \subsection{Coverage-Aware Grid Query (CAGQ)}
        In this subsection, we discuss the details of the CAGQ module. Given a point cloud, CAGQ aims to effectively structure the point cloud, and ease the process of center sampling and neighbor points querying. To perform CAGQ, we first voxelize the input space by setting up a voxel size $(v_x, v_y, v_z)$. We then map each point to a voxel index $Vid(u,v,w) = floor(\frac{x}{v_x}, \frac{y}{v_y}, \frac{z}{v_z})$. Here we only store up to $n_v$ points in each voxel. 
        
        Let $O_v$ denote all of the non-empty voxels. We then sample $M$ \textit{center voxels} $O_c \subseteq O_v$. For each center voxel $v_i$, we define its voxel neighbors $\pi(v_i)$ as the voxels within the neighbor-hood of a center voxel. In Figure 2d, $\pi(v(2,1))$ are the 3X3 voxels inside the red box. We call the stored points inside $\pi(v_i)$ as \textit{context points}. Since we build the point-voxel index in the previous step, CAGQ can quickly retrieve context points for each $v_i$.
        
        After that, CAGQ picks $K$ \textit{node points} from the context points of each $v_i$. We calculate the barycenter of node points in a group, as the location of the group center. This entire process is shown in Figure \ref{fig:CAGQ}.
        
        Two problems remain to be solved here. (1) How do we sample center voxels $O_c \subseteq O_v$. (2) How do we pick $K$ nodes from context points in $\pi(v_i)$.
        
        To solve the first problem, we propose our \textbf{center voxels sampling} framework, which includes two methods:
        
        1. \textit{Random Voxel Sampling (RVS)}: Each occupied voxel will have the same probability of being picked. The group centers calculated inside these center voxels are more evenly distributed than centers picked on input points by RPS. We discuss the details in Section \ref{sec:anacagq}.
        
        2. \textit{Coverage-Aware Sampling (CAS)}: Each selected center voxel can cover up to $\lambda$ occupied voxel neighbors. The goal of CAS is to select a set of center voxels $O_c$ such that they can cover the most occupied space. Seeking the optimal solution to this problem requires iterating all combinations of selections. Therefore, we employ a greedy algorithm to approach the optimal solution: We first randomly pick $M$ voxels from $O_v$ as \textit{incumbents}; From all of the unpicked voxels, we iteratively select one to challenge a random incumbent each time. If adding this \textit{challenger} (and in the meantime removes the incumbent) gives us better coverage, we replace the incumbent with the challenger. For a challenger $v_C$ and an incumbent $v_I$, the heuristics are calculated as:
        \begin{adjustwidth}{0pt}{0pt}
            \setlength{\abovedisplayskip}{-10pt}%
            \setlength{\abovedisplayshortskip}{\abovedisplayskip}%
            \setlength{\belowdisplayskip}{5pt}%
            \begin{align}
            &\delta(x) =
            \begin{cases}
            1, & \text{if $x=0$}.\\
            0, & \text{otherwise}.
            \end{cases} \\
            &H_{add} = \sum_{V \in \pi(V_{C})}{ \delta(C_V) - \beta \cdot \cfrac{C_V}{\lambda} } \label{eq:hadd} \\        
            &H_{rmv} = \sum_{V \in \pi(V_{I})}{ \delta(C_V-1)} \label{eq:hrmv}              
            \end{align}
        \end{adjustwidth}
        where $\lambda$ is the amount of neighbors of a voxel and $C_V$ is the number of incumbents covering voxel $V$. $H_{add}$ represents the coverage gain if adding $V_C$ (penalize by a term of over-coverage). $H_{rmv}$ represents the coverage loss after removing $V_I$. If $H_{add} > H_{rmv}$, we replace the incumbent by the challenger voxel. If we set $\beta$ as 0, each replacement is guaranteed to improve the space coverage.
    
        Comparisons of those methods are further discussed in section \ref{sec:anacagq}.
        
        \textbf{Node points querying} CAGQ also provides two strategies to pick $K$ node points from context points in $\pi(v_i)$.
        
        1. \textit{Cube Query}: We randomly select $K$ points from context points. Compared to the Ball Query used in PointNet++ \cite{qi2017pointnet++}, Cube Query can cover more space when point density is imbalanced. In the scenario of Figure \ref{fig:CAGQ}, Ball Query samples $K$ points from all raw points (grey) and may never sample any node point from voxel (2,1) which only has 3 raw points.
        
        2. \textit{K-Nearest Neighbors}: Unlike traditional k-NN where the search space is all points, k-NN in CAGQ only need to search among the context points, making the query substantially faster (We also provide an optimized method in the supplementary materials). We will compare these methods in the next section. 
        
    \subsection{Grid Context Aggregation}
    For each point group provided by CAGQ, we use a Grid Context Aggregation (GCA) module to aggregate features from the node points to the group center. We first construct a local graph $G(V, E)$, where $V$ consists of the group center and $K$ node points provided by CAGQ. We then connect each node point to the group center. GCA projects a node point's features $f_i$ to $\widetilde{f}_i$. Based on the edge relation between the node and the center, GCA calculates the contribution of $\widetilde{f}_i$ and aggregates all these features as the feature of the center $\widetilde{f_c}$. Formally, the GCA module can be described as
    \begin{adjustwidth}{0pt}{0pt}
        \setlength{\abovedisplayskip}{-5pt}%
        \setlength{\abovedisplayshortskip}{\abovedisplayskip}%
        \setlength{\belowdisplayskip}{0pt}%
        \begin{align}
        \widetilde{f}_{c,i} = e(\chi_i, f_i) * \mathcal{M}(f_i)\\
        \widetilde{f_c} = \mathcal{A}(\{\widetilde{f}_{c,i}\}, i \in {1,...,K})
        \end{align}
    \end{adjustwidth}
    where $\widetilde{f}_{c,i}$ is the contribution from a node, and $\chi_i$ is the xyz location of the node. $\mathcal{M}$ is a multi-layer perceptron (MLP), $e$ is the edge attention function, and $\mathcal{A}$ is the aggregation function. The edge attention function $e$ has been explored by many previous studies \cite{xu2018spidercnn,cciccek20163d,wang2019graph}. In this work, we design a new edge attention function with the following improvements to better fit into our network architecture (Figure \ref{fig:edgefeatnet}):
    \begin{figure}[!t]
        \setlength{\abovedisplayskip}{0pt}%
        \setlength{\abovedisplayshortskip}{\abovedisplayskip}%
        \begin{adjustwidth}{0pt}{0pt}
            \centering
            \includegraphics[width=0.7\linewidth]{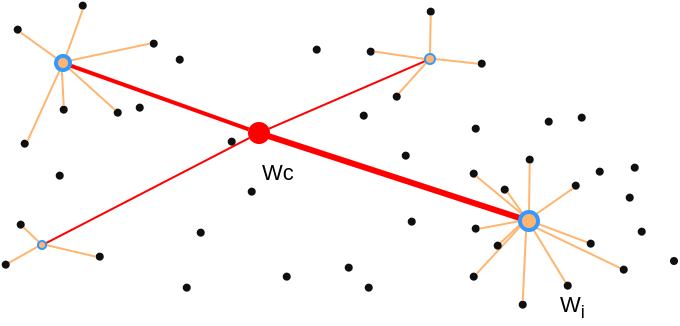}
            \captionsetup{aboveskip=5pt}
            \captionsetup{belowskip=-10pt}
            \caption{The red point is the group center. Yellow points are its node points. Black points are node points of the yellow points in the previous layer. The coverage weight is an important feature as it encodes the number of black points that have been aggregated to each yellow point.}
            \label{fig:cw}
        \end{adjustwidth}
    \end{figure}
    
    \textbf{Coverage Weight} Previous studies \cite{xu2018spidercnn,liu2019relation,lan2019modeling,wang2019dynamic,zhang2019linked} use $\chi_c$ of the center and $\chi_i$ of a node to model edge attention as a function of geometric relation (Figure \ref{fig:edgefeatnet}b). However, the formulation ignores the underlying contribution of each node point from previous layers. Intuitively, node points with more information from previous layers should be given more attention. We illustrate this scenario in Figure \ref{fig:cw}. With that in mind, we introduce the concept of \textit{coverage weight}, which is defined as the number of points that have been aggregated to a node in previous layers. This value can be easily computed in CAGQ, and we argue that coverage weight is an important feature in calculating edge attention (see our ablation studies in Table \ref{tab:ablation}).
    
    \begin{figure*}[!hbt]
        \begin{adjustwidth}{-12pt}{0pt}
            \centering
            \includegraphics[width=1.03\linewidth]{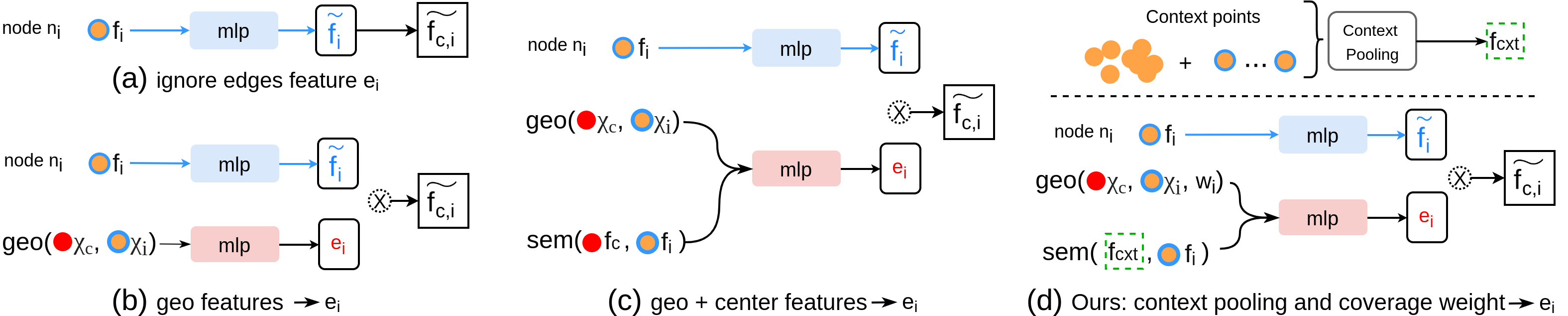}
            \captionsetup{aboveskip=5pt}
            \captionsetup{belowskip=-10pt}
            \caption{Different strategies to compute the contribution $\widetilde{f_{c,i}}$ from a node $n_i$ to its center $c$. $f_i, \chi_i$ are the feature maps and the location of $n_i$. $e_i$ is the edge feature between $n_i$ and $c$ calculated from the edge attention function. \textbf{(a)} Pointnet++ \cite{qi2017pointnet++} ignores $e_i$. \textbf{(b)} computes $e_i$ based on low dimensional geometric relation between $n_i$ and $c$. \textbf{(c)} also consider semantic relation between the center and the node point, but $c$ has to be sampled on one of the points from the previous layer. \textbf{(d)}. Grid-GCN's geo-relation also includes the coverage weight. It pools a context feature $f_{cxt}$ from all stored neighbors to provide a semantic reference in $e_i$ computing.}
            \label{fig:edgefeatnet}
        \end{adjustwidth}
    \end{figure*}
    
    \textbf{Grid Context Pooling} Semantic relation is another important aspect when calculating the edge attention. In previous works \cite{cciccek20163d,wang2019graph}, semantic relation is encoded by using the group center's features $f_c$ and a node point's features $f_i$, which requires the group center to be selected from node points. In CAGQ, since a group center is calculated as the barycenter of the node points, we propose \textit{Grid context pooling} that extracts context features $f_{cxt}$ by pooling from all context points, which sufficiently covers the entire grid space of the local graph. Grid context pooling brings the following benefits: 
    \begin{itemize}[itemsep=-1pt,topsep=0pt,leftmargin=15pt]
        \item $f_{cxt}$ models the features of a virtual group center, which allows us to calculate the semantic relation between the center and its node points.
        \item Even when group center is picked on a physical point, $f_{cxt}$ is still a useful feature representation as it covers more points in the neighborhood, instead of only the points in the graph.
        \item Since we have already associated context points to its center voxel in CAGQ, there is no extra point query overhead. $f_{cxt}$ is shared across all edge computation in a local graph, and the pooling is a light-weighted operation requiring no learnable weights, which introduces little computational overhead.
    \end{itemize}
    GCA module is summarized in Figure \ref{fig:edgefeatnet}d, and the edge attention function can be model as 
    \begin{adjustwidth}{0pt}{0pt}
        \setlength{\abovedisplayskip}{-5pt}%
        \setlength{\abovedisplayshortskip}{\abovedisplayskip}%
        \setlength{\belowdisplayskip}{0pt}%
        \begin{align}
        e = mlp(mlp_{geo}(\chi_c, \chi_i, w_i), mlp_{sem}(f_{cxt}, f_i))
        \end{align}
    \end{adjustwidth}    
\section{Analysis of CAGQ}
    \label{sec:anacagq}
        \begin{figure*}[!hbt]
            \setlength{\abovedisplayskip}{0pt}%
            \setlength{\abovedisplayshortskip}{\abovedisplayskip}%
            \setlength{\belowdisplayskip}{5pt}%
            \vspace{-10pt}
            \begin{adjustwidth}{-10pt}{-10pt}
            
                \begin{subfigure}{0.33\linewidth}
                    \centering
                    \includegraphics[width=1\linewidth]{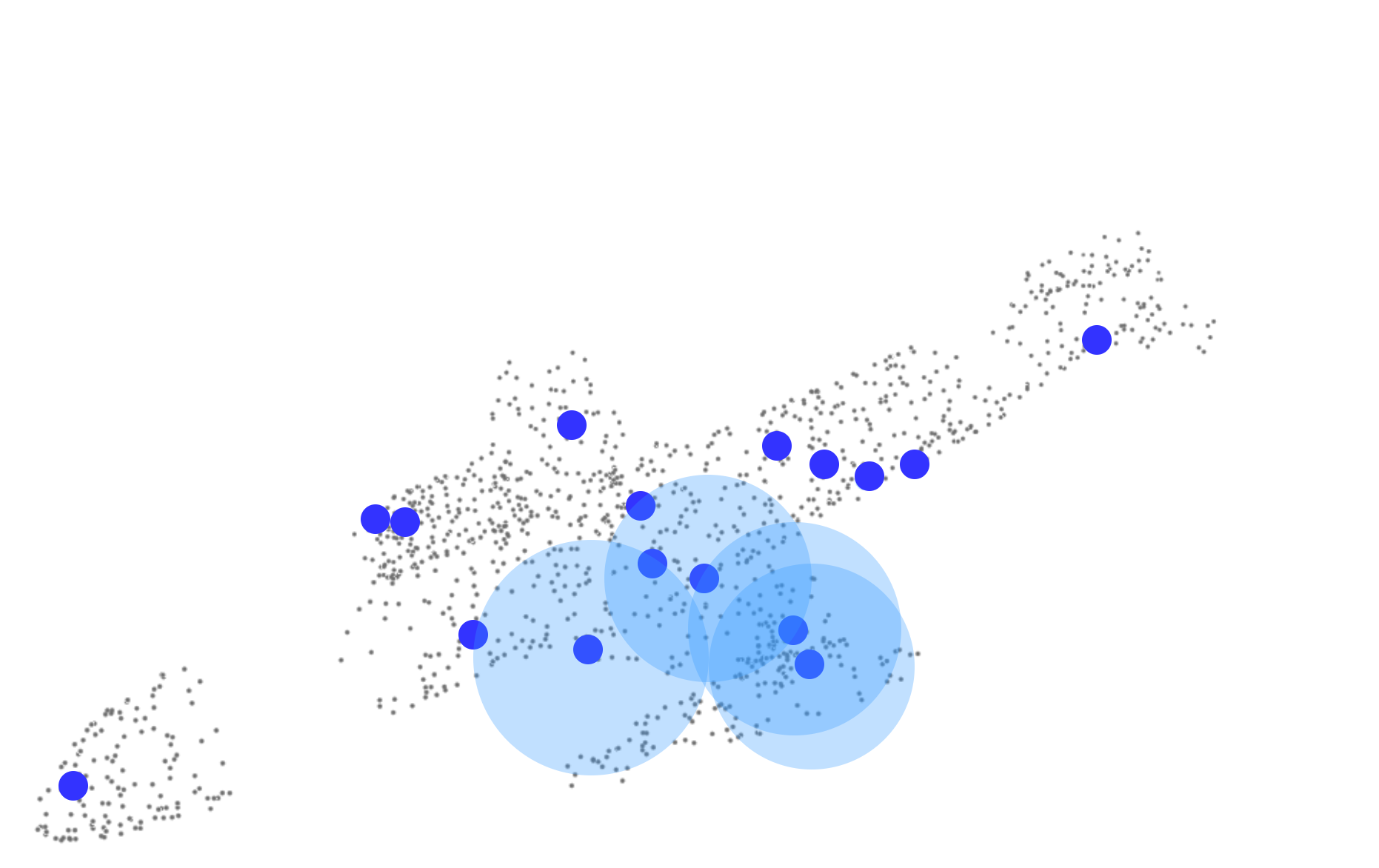}
                    \caption{Random Point Sampling}
                \end{subfigure}
                \begin{subfigure}{0.33\linewidth}
                    \centering
                    \includegraphics[width=1\linewidth]{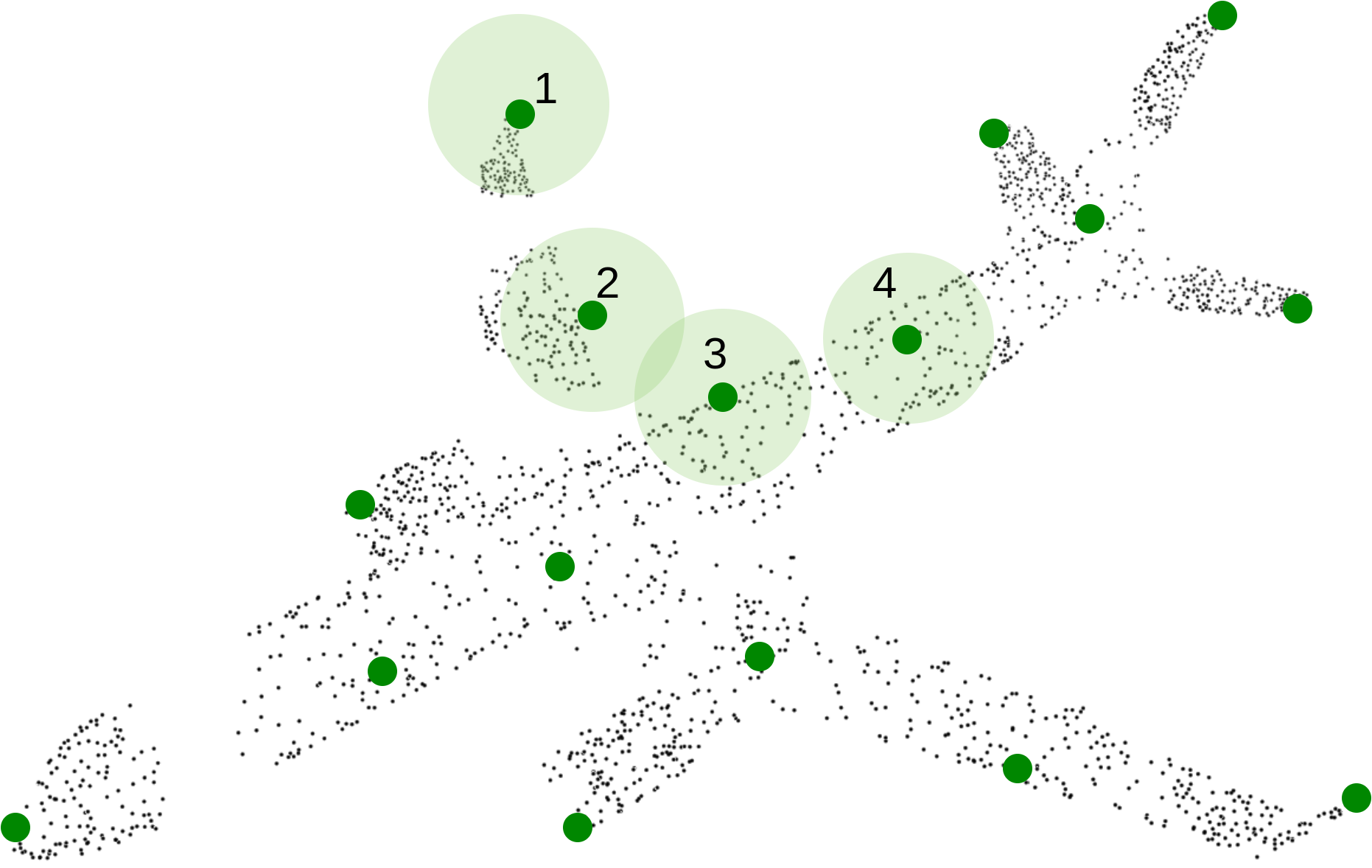}
                    \caption{Farthest Point Sampling}
                \end{subfigure}
                \begin{subfigure}{0.33\linewidth}
                    \centering
                    \includegraphics[width=1\linewidth]{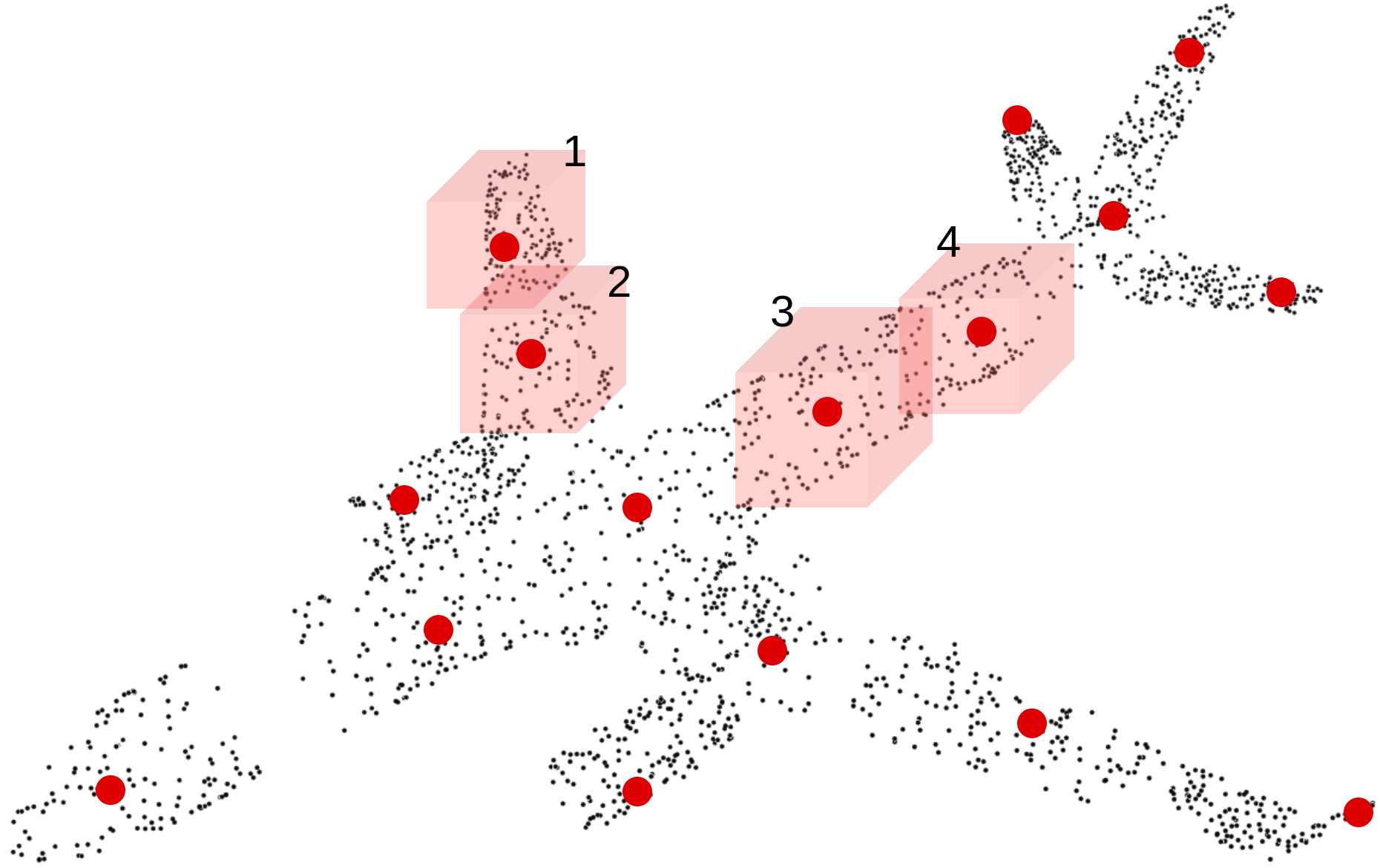}
                    \caption{Coverage-Aware Sampling}
                \end{subfigure}
            \captionsetup{belowskip=-10pt}
            \caption{The visualization of the sampled group center and the queried node points by RPS, FPS, and CAS. The blue and green balls indicate Ball Query. The red squares indicate Cube Query. The ball and cube have the same volume. (a) RPS covers $45.6 \%$  of the occupied space,  while FPS covers $65 \%$ and CAS covers $75.2 \%$.}
            \label{fig:sampling_qual}
        \end{adjustwidth}
    \end{figure*}
    To analyze the benefit of CAGQ, we test the occupied space coverage and the latency of different sampling/querying methods under different conditions on ModelNet40 \cite{wu20153d}. Center sampling methods include Random Point Sampling (RPS), Farthest Point Sampling (FPS), our Random Voxel Sampling (RVS), and our Coverage-Aware Sampling (CAS). Neighbor querying methods include Ball Query, Cube query, and K-Nearest Neighbors. The conditions include different numbers of input points, node numbers in a point group, and numbers of point groups, which are denoted by $N$, $K$, and $M$. We summarize the qualitative and quantitative evaluation result in Table \ref{tab:query} and Figure \ref{fig:sampling_qual}. 
    The reported occupied space coverage is calculated as the ratio between the number of voxels occupied by node points of all groups, and the number of voxels occupied by the original $N$ points. Results under more conditions are presented in the supplementary.
    \subsection{Space Coverage}
     In Figure \ref{fig:sampling_qual}a, the centers sampled by RPS are concentrated in the areas with higher point density, leaving most space uncovered. In Figure \ref{fig:sampling_qual}b, FPS picks the points that are far away from each other, mostly on the edges of the 3D shape, which causes the gap between centers. In Figure \ref{fig:sampling_qual}c, our CAS optimizes the voxel selection and covers $75.2\%$ of occupied space. Table \ref{tab:query} lists the percentage of space coverage by RPS, FPS, RVS, and CAS. CAS leads the space coverage in all cases (30 \% more than RPS). FPS has no advantage over RVS when $K$ is small. 
    
    The factors that benefit CAGQ in space coverage can be summarized as follows:
    \begin{itemize}[itemsep=-1pt,topsep=3pt,leftmargin=15pt]
        \item Instead of sampling centers from $N$ points, RVS samples center voxels from occupied space, therefore it is more resilient to point density imbalance (Figure \ref{fig:sampling_qual}).
        \item CAS further optimizes the result of RVS by conducting a greedy candidate replacement. Each replacement is guaranteed to result in better coverage.
        \item CAGQ stores the same number of points in each occupied voxel. The context points are more evenly distributed, so are the $K$ node points picked from the context points. Consequently, the strategy reduces the coverage loss caused by density imbalance in a local area.
    \end{itemize}
    \subsection{Time complexity}
    
    We summarize the time complexity of different methods in Table \ref{tab:complexity}. The detailed deduction is presented in the supplementary. Table \ref{tab:query} shows the empirical results of latency. We see that our CAS is much faster than FPS and achieves $50\times$ speed-up. CAS + Cube Query can even outperform RPS + Ball Query when the size of the input point cloud is large. This is due to the higher neighborhood query speed. Because of better time complexity, RVS + k-NN leads the performance under all conditions and achieves $6\times$ speed-up over FPS + k-NN.
    
    \begin{table}[!hbt]
        \setlength{\abovedisplayskip}{0pt}%
        \setlength{\abovedisplayshortskip}{\abovedisplayskip}%
        \setlength{\belowdisplayskip}{5pt}%
        \begin{adjustwidth}{-5pt}{0pt}
            \setlength\tabcolsep{1.5pt}
            \begin{tabular}{c|cccc}
                \hline
                \multirow{2}{*}{\begin{tabular}[c]{@{}c@{}}Sample \\ centers\end{tabular}}  & RPS        & FPS\cite{eldar1997farthest}        & RVS*       & CAS*      \\
                & $O(N)$ & $O(NlogN)$    & $O(N)$ & $O(N)$    \\ \hline
                \multirow{2}{*}{\begin{tabular}[c]{@{}c@{}}Query \\ nodes\end{tabular}} & Ball Query & Cube Query* & k-NN\cite{cover1967nearest}       & CAGQ k-NN* \\
                & $O(MN)$    & $O(MK)$     & $O(MN)$   & $O(Mn_v)$ \\ \hline
            \end{tabular}
            \captionsetup{aboveskip= 5pt}
            \captionsetup{belowskip=-10pt}
            \caption{Time complexity: We sample $M$ centers from $N$ points and query $K$ neighbors per center. We limit the maximum number of points in each voxel to $n_v$. In practice, $K < N$, and $n_v$ is usually of the same magnitude to $K$. Approximate FPS algorithm can be $O(NlogN)$\cite{eldar1992irregular}. * indicates our methods. See the supplementary for deduction details.}
            \label{tab:complexity}
        \end{adjustwidth}
    \end{table}
    \begin{table*}[]
        \begin{adjustwidth}{-10pt}{0pt}
            \setlength\tabcolsep{5pt}
            \begin{tabular}{ccc|cccc|cccccccc}
                \hline
                 \multicolumn{3}{c|}{Center sampling} & RPS  & FPS            & RVS*           & CVS*           & RPS           & FPS    & RVS*          & \multicolumn{1}{c|}{CVS*}  & RPS    & FPS    & RVS*           & CVS*  \\ \hline
                 \multicolumn{3}{c|}{Neighbor querying}  & Ball & Ball           & Cube           & Cube           & Ball          & Ball   & Cube          & \multicolumn{1}{c|}{Cube}  & k-NN    & k-NN    & k-NN            & k-NN   \\ \hline
                N                      & K                & M                 & \multicolumn{4}{c|}{Occupied space coverage(\%)}        & \multicolumn{8}{c}{Latency (ms) with batch size = 1}                                                           \\ \hline
                \multirow{4}{*}{1024}  & 8                & 8                 & 12.3 & 12.9           & 13.1           & \textbf{14.9}  & \textbf{0.29} & 0.50   & 0.51          & \multicolumn{1}{c|}{0.74}  & 0.84   & 0.85   & \textbf{0.51}  & 0.77  \\
                & 8                & 128               & 64.0 & 72.5           & 82.3           & \textbf{85.6}  & \textbf{0.32} & 0.78   & 0.44          & \multicolumn{1}{c|}{0.68}  & 1.47   & 1.74   & \textbf{0.52}  & 0.72  \\
                & 128              & 32                & 60.0 & 70.1           & 61.0           & \textbf{74.7}  & \textbf{0.37} & 0.53   & 0.96          & \multicolumn{1}{c|}{1.18}  & 22.23  & 21.08  & \textbf{2.24}  & 2.74  \\
                & 128              & 128               & 93.6 & 99.5           & 95.8           & \textbf{99.7}  & \textbf{0.38} & 0.69   & 1.03          & \multicolumn{1}{c|}{1.17}  & 32.48  & 32.54  & \textbf{6.85}  & 7.24  \\ \hline
                \multirow{4}{*}{8192}  & 8                & 64                & 19.2 & 22.9           & 22.1           & \textbf{25.1}  & \textbf{0.64} & 1.16   & 0.66          & \multicolumn{1}{c|}{0.82}  & 1.58   & 1.80   & \textbf{0.65}  & 0.76  \\
                & 8                & 1024              & 82.9 & \textbf{96.8}  & 92.4           & 94.4           & 0.81          & 4.90   & \textbf{0.54} & \multicolumn{1}{c|}{0.87}  & 1.53   & 5.36   & \textbf{0.93}  & 0.97  \\
                & 128              & 256               & 79.9 & 90.7           & 80.0           & \textbf{93.5}  & 1.19          & 1.19   & \textbf{1.17} & \multicolumn{1}{c|}{1.41}  & 21.5   & 21.5   & \textbf{15.19} & 17.68 \\
                & 128              & 1024              & 98.8 & 99.9           & 99.5           & \textbf{100.0} & \textbf{1.22} & 5.25   & 1.40          & \multicolumn{1}{c|}{1.76}  & 111.4  & 111.7  & \textbf{24.18} & 27.65 \\ \hline
                \multirow{4}{*}{81920} & 32               & 1024              & 70.6 & 86.3           & 78.3           & \textbf{91.6}  & 8.30          & 33.52  & \textbf{3.34} & \multicolumn{1}{c|}{6.02}  & 19.49  & 43.69  & \textbf{8.76}  & 10.05 \\
                & 32               & 10240             & 98.8 & 99.2           & \textbf{100.0} & \textbf{100.0} & 8.93          & 260.48 & \textbf{4.22} & \multicolumn{1}{c|}{9.35}  & 20.38  & 272.48 & \textbf{9.65}  & 17.44 \\
                & 128              & 1024              & 72.7 & 88.2           & 79.1           & \textbf{92.6}  & 9.68          & 34.72  & \textbf{4.32} & \multicolumn{1}{c|}{8.71}  & 71.99  & 93.02  & \textbf{50.7}  & 61.94 \\
                & 128              & 10240             & 99.7 & \textbf{100.0} & \textbf{100.0} & \textbf{100.0} & 10.73         & 258.49 & \textbf{5.83} & \multicolumn{1}{c|}{11.72} & 234.19 & 442.87 & \textbf{69.02} & 83.32 \\ \hline
            \end{tabular}
            \captionsetup{aboveskip=5pt}
            \captionsetup{belowskip=-10pt}
            \caption{Performance comparisons of data structuring methods, run on ModelNet40\cite{wu20153d}. Center sampling methods include RPS, FPS, CAGQ's RVS and CAS. Neighbor querying methods include Ball Query, Cube query and K-Nearest Neighbors. Condition variables include N points, M groups and K neighbors per group. Occupied space coverage = num. of occupied voxels of queried points / num. of occupied voxels of the original N points.}
            \label{tab:query}
        \end{adjustwidth}
    \end{table*}
    
\section{Experiments}

    \begin{table}[ht!]
        \begin{adjustwidth}{-10pt}{0pt}
            \setlength\tabcolsep{3pt}
            \begin{tabular}{ll|cccc|c}
                \hline
                &       & \multicolumn{2}{c}{ModelNet40} & \multicolumn{2}{c|}{ModelNet10} & \multirow{2}{*}{\begin{tabular}[c]{@{}c@{}}latency\\ (ms)\end{tabular}} \\ \cline{1-6}
                \multicolumn{2}{c}{Input (xyz as default)}  & OA            & mAcc           & OA            & mAcc            &                                                                         \\ \hline \hline
                \multicolumn{7}{c}{ OA $\leqslant 91.5$}  \\ 
                PointNet\cite{qi2017pointnet}              & 16$\times$1024       & 89.2            & 86.2           & -               & -              & \textbf{15.0}                                                                     \\
                SCNet\cite{xie2018attentional}             &  16$\times$1024       & 90.0            & 87.6           & -               &                & -                                                                        \\
                SpiderCNN\cite{xu2018spidercnn}             &  8   $\times$ 1024       & 90.5            & -              & -               & -              & 85.0                                                                     \\
                O-CNN\cite{Wang-2017-ocnn}                 & octree   & 90.6            & -              & -               & -              &     90.0                                                                     \\
                SO-net\cite{li2018so}                &         8   $\times$ 2048       & 90.8            & 87.3           & \textbf{94.1}   & \textbf{93.9}  & -                                                                        \\
                \textbf{Grid-GCN$^1$}         & 16$\times$1024       & \textbf{91.5}   & \textbf{88.6}  & 93.4            & 92.1           & 15.9                                                            \\ \hline \hline
                \multicolumn{7}{c}{ OA $\leqslant 92.0$}  \\ 
                3DmFVNet\cite{ben20183dmfv}       & 16$\times$1024       & 91.6            & -              & 95.2            & -              & 39.0                                                                     \\
                PAT\cite{yang2019modeling}                   &  8  $\times$ 1024       & 91.7            & -              &                 & -              & 88.6                                                                     \\
                Kd-net\cite{klokov2017escape}                & kd-tree    & 91.8            & 88.5           & 94.0            & 93.5           & -                                                                        \\
                PointNet++\cite{qi2017pointnet++}            & 16$\times$1024       & 91.9            & \textbf{90.7}  &                 & -              & 26.8                                                                     \\
                \textbf{Grid-GCN$^2$}  & 16$\times$1024       & \textbf{92.0}   & 89.7           & \textbf{95.8}   & \textbf{95.3}  & \textbf{21.8}                                                            \\ \hline \hline
                \multicolumn{7}{c}{ OA $> 92.0$}  \\ 
                DGCNN\cite{wang2019dynamic}              & 16$\times$1024       & 92.2            & 90.2           &                 & -              & 89.7                                                                     \\
                PCNN\cite{atzmon2018point}                  & 16$\times$1024       & 92.3            & -              & 94.9            & -              & 226.0                                                                    \\
                Point2Seq\cite{liu2019point2sequence}             & 16$\times$1024       & 92.6            & -              &                 &                & -                                                                        \\
                A-CNN\cite{komarichev2019cnn}                  & 16$\times$1024  & 92.6            & 90.3           & 95.5            & 95.3           & 68.0                                                                     \\
                KPConv\cite{thomas2019kpconv}                & 16$\times$6500       & 92.7            & -              & -               & -              & 125.0                                                                    \\
                \textbf{Grid-GCN$^3$}  & 16$\times$1024       & 92.7            & 90.6           & 96.5            & 95.7           & \textbf{26.2}                                                            \\
                \textbf{Grid-GCN$^{full}$}  & 16$\times$1024       & \textbf{93.1}   & \textbf{91.3}  & \textbf{97.5}   & \textbf{97.4}  & 42.2                                                                     \\ \hline
            \end{tabular}
            \captionsetup{aboveskip=5pt}
            \captionsetup{belowskip=-20pt}
            \caption{Results on ModelNet10 and ModelNet40\cite{wu20153d}. Our full model achieves the state-of-the-art accuracy. With model reduction, our compact models Grid-GCN$^{1-3}$ also out speed other models. We discuss their details in the ablation studies.}
            \label{tab:cls}
        \end{adjustwidth}
    \end{table}
    
    We evaluate Grid-GCN on multiple datasets: ModelNet10 and ModelNet40\cite{wu20153d} for object classification, ScanNet\cite{dai2017scannet} and S3DIS\cite{2017arXiv170201105A} for semantic segmentation. Following the convention of PVCNN \cite{liu2019point}, we report latency and performance in each level of accuracy. We collect the result of other models either from published papers or the authors. All the latency results are reported under the corresponding batch size and number of input points. All experiments are conducted on a single RTX 2080 GPU. Training details are listed in the supplementary. 
    
    \subsection{3D Object Classification}
    
    \textbf{Datasets and settings} We conduct the classification tasks on the ModelNet10 and ModelNet40 dataset\cite{wu20153d}. ModelNet10 is composed of 10 object classes with 3991 training and 908 testing objects. ModelNet40 includes 40 different classes with 9843 training objects and 2468 testing objects. We prepare our data following the convention of PointNet\cite{qi2017pointnet}, which uses 1024 points with 3 channels of spatial location as input. Several studies use normal \cite{qi2017pointnet++,komarichev2019cnn}, octree \cite{Wang-2017-ocnn}, or kd-tree for input, and \cite{liu2019relation,liu2019densepoint} use voting for evaluation. 
    
    \textbf{Evaluation} To compare with different models with different levels of accuracy and speed, we train Grid-GCN with 4 different settings to balance performance and speed (Details are shown in section \ref{ablation}). The variants are in the number of feature channels and the number of node points in a group in the first layer (see Table \ref{tab:ablation}). The results are shown in Table \ref{tab:cls}. We report our results without voting. For all of the four settings, our Grid-GCN model not only achieves state-of-the-art performance on both ModelNet10 and ModelNet40 datasets, but has the best speed-accuracy trade-off. Although Grid-GCN uses the CAGQ module for data structuring, it has similar latency as PointNet which has no data structuring step while its accuracy is significantly higher than PointNet. 
    
    \subsection{3D Scene Segmentation}
        \textbf{Dataset and Settings} We evaluate our Grid-GCN on two large-scale point cloud segmentation datasets: ScanNet\cite{dai2017scannet} and Stanford 3D Large-Scale Indoor Spaces (S3DIS) \cite{2017arXiv170201105A}. ScanNet consists of 1513 scanned indoor scene, and each voxel is annotated in 21 categories. We follow the experiment setting in \cite{dai2017scannet} and use 1201 scenes for training, and 312 scenes for testing. Following the routine and evaluation protocol in PointNet++\cite{qi2017pointnet++}, we sample 8192 points during training and 3 spatial channels for each point. S3DIS contains 6 large-scale indoor areas with 271 rooms. Each point is labeled with one of 13 categories. Since area 5 is the only area that doesn't have overlaps with other areas, we follow \cite{tchapmi2017segcloud,li2018pointcnn,liu2019point} to train on area 1-4 and 6, and test on area 5. In each divided section, 4096 points are sampled for training, and we adopt the evaluation method from \cite{li2018pointcnn}. 
        
        \textbf{Evaluation} We report the overall voxel labeling accuracy (OA) and the runtime latency for ScanNet\cite{dai2017scannet}. We trained two versions of the Grid-GCN model, with a full model using $1\times K$ node points and a compact model using $0.5\times K$ node points. Results on are reported in Table \ref{tab:scannet}. 
        
        Since the segmentation tasks generally use more input points than the classification model, our advantage of data structuring becomes outstanding. With the same amount of input points (32768) in a batch, Grid-GCN out-speed PointNet++ $4.5\times$ while maintaining the same level of accuracy. Compared with more sophisticated models such as PointCNN \cite{li2018pointcnn} and A-CNN \cite{komarichev2019cnn}, Grid-GCN is $25\times$ and $12\times$ faster, respectively, while achieving the state-of-the-art accuracy. Remarkably, Grid-GCN can run as fast as \textbf{50} to \textbf{133} FPS with state-of-the-art performance, which is desired in real-time applications. A popular model MinkowskiNet\cite{choy20194d} doesn't report the overall accuracy, therefore we don't put it in the table. But its github example shows a latency of 103ms on Scannet.
        
\begin{figure}[!t]
                \setlength{\abovedisplayskip}{0pt}%
                \setlength{\abovedisplayshortskip}{\abovedisplayskip}%
                \setlength{\belowdisplayskip}{5pt}%
                \begin{adjustwidth}{-5pt}{-2pt}
                    \centering
                    \begin{subfigure}{0.48\linewidth}
                        \centering
                        \includegraphics[width=1\linewidth]{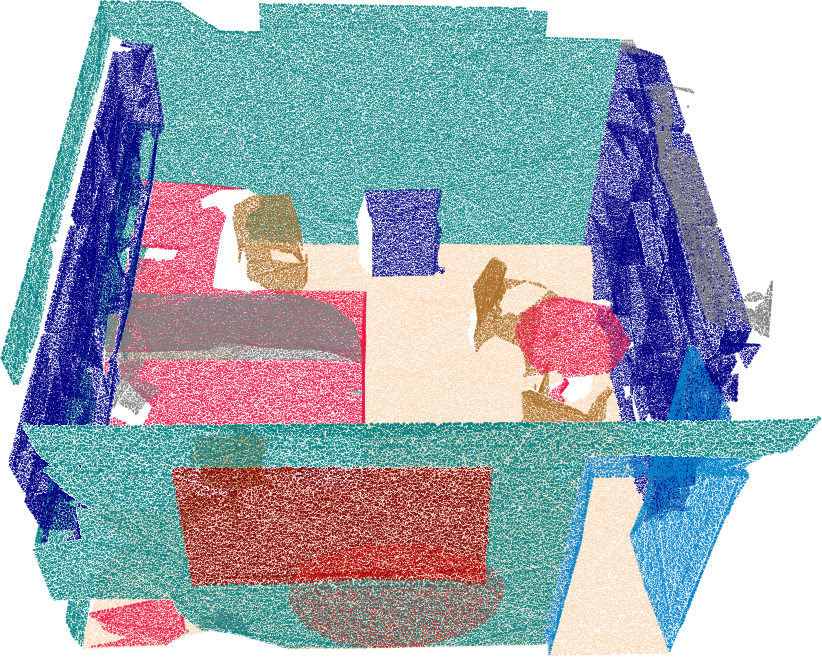}
                    \end{subfigure} \hfill
                    \begin{subfigure}{0.48\linewidth}
                        \centering
                    \includegraphics[width=1\linewidth]{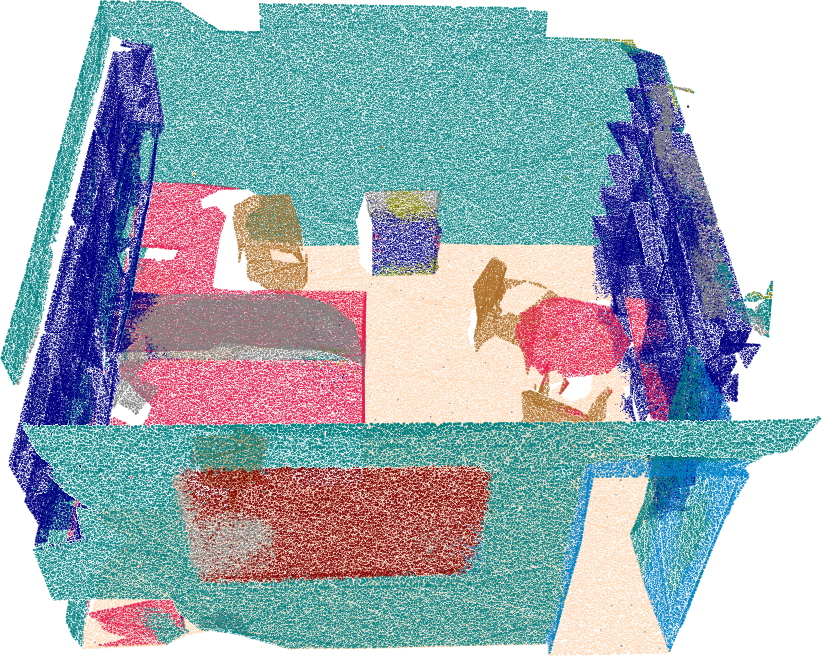}
                    \end{subfigure}
                    \begin{subfigure}{0.48\linewidth}
                        \centering
                        \includegraphics[width=1\linewidth]{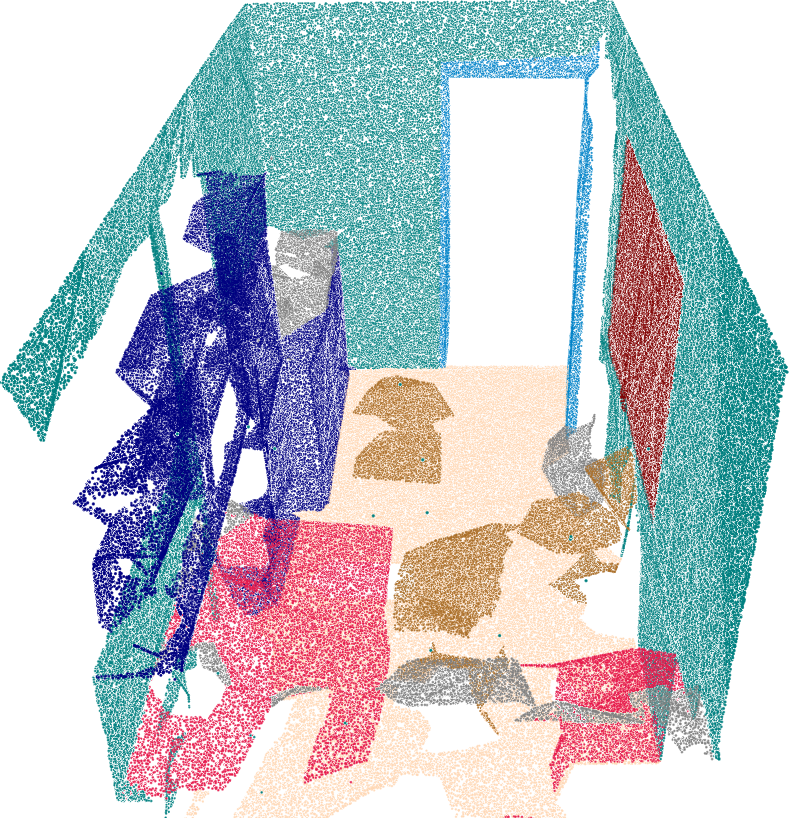}
                        \caption{Ground Truth}
                        \label{fig:visual_gt}
                    \end{subfigure} \hfill
                    \begin{subfigure}{0.48\linewidth}
                        \centering
                    \includegraphics[width=1\linewidth]{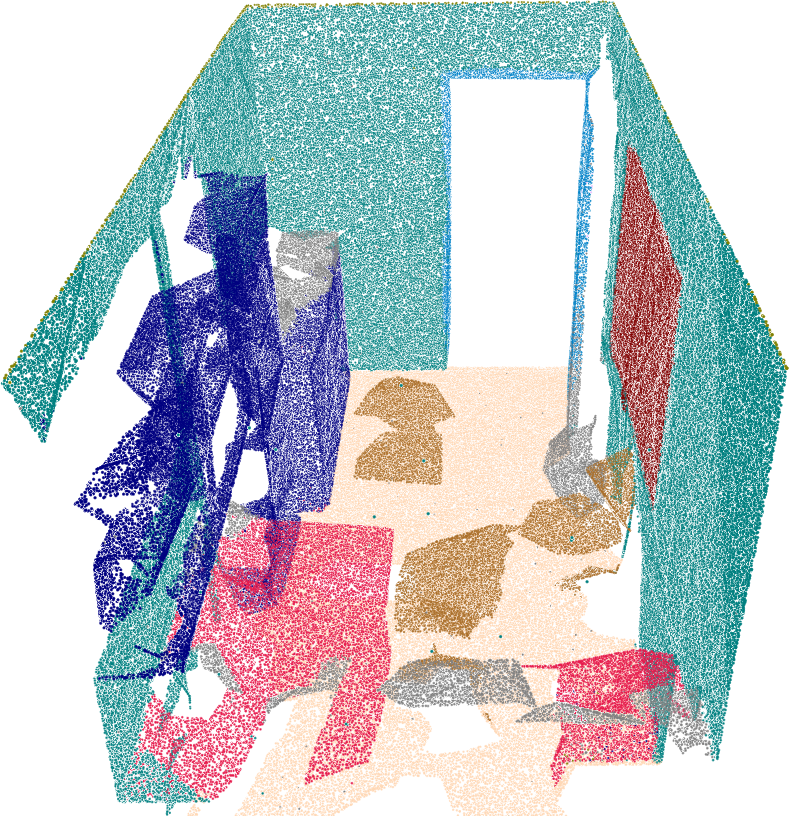}
                        \caption{Ours}
                        \label{fig:visual_ours}
                    \end{subfigure}
                    \captionsetup{aboveskip=5pt}
                    \captionsetup{belowskip=0pt}
                    \caption{Semantic segmentation results on S3DIS \cite{2017arXiv170201105A} area 5.}
                    \label{fig:S3DIS}
                \end{adjustwidth}
            \end{figure}

        We show the quantitative results on S3DIS in Table \ref{tab:s3dis} and visual result in Figure \ref{fig:S3DIS}. Our compact version of Grid-GCN is generally $4\times$ to $14\times$ faster than other models with data structuring. Notably, even compared with PointNet that has no data structuring at all, we are still $1.6\times$ faster while achieves $12\%$ performance gain in mIOU. For our full model, we are still the fastest and achieve $2\times$ speed-up over PVCNN++\cite{liu2019point}, a state-of-the-art study focusing on speed improvement.

            \begin{table}[t!]
                \begin{adjustwidth}{5pt}{-5pt}
                    \begin{tabular}{ll@{\hskip20pt}c@{\hskip5pt}c}
                        \hline
                        \multicolumn{2}{c}{Input (xyz as default)}&  OA & latency (ms) \\ \cline{1-1} \hline
                        \multicolumn{4}{c}{OA $< 84.0$}  \\ 
                        PointNet\cite{qi2017pointnet}                                & 8 $\times$ 4096                                                  & 73.9                                                  & 20.3                                                    \\
                        OctNet\cite{riegler2017octnet}                                  & volume                       & 76.6                                              &  -                                                       \\
                        PointNet++\cite{qi2017pointnet++}                     & 8 $\times$ 4096                                                  & 83.7                                                  & 72.3                                                    \\
                        \textbf{Grid-GCN}$_{(0.5 \times K)}$                 & 4 $\times$ 8192                                                  & \textbf{83.9}                                         & \textbf{16.6}                                           \\ \hline \hline
                        \multicolumn{4}{c}{ OA $\geqslant 84.0$}  \\ 
                        SpecGCN\cite{wang2018local}                                 & -                                                  & 84.8                                                  & -                                                       \\
                        PointCNN\cite{li2018pointcnn}                               & 12$\times$2048                                                  & 85.1                                                  & 250.0                                                   \\
                        Shellnet\cite{zhang2019shellnet}                                & -                                                     & 85.2                                                  & -                                                       \\
                        \textbf{Grid-GCN}$_{(1 \times K)}$                   & 4 $\times$ 8192                                                  & \textbf{85.4}    & \textbf{20.8}                                           \\ \hline
                        A-CNN\cite{komarichev2019cnn}                                    & 1 $\times$ 8192                                                   & \textbf{85.4}                                                  & 92.0                                                    \\
                        \textbf{Grid-GCN}$_{(1 \times K)}$                   & 1 $\times$ 8192                                                  & \textbf{85.4}                                         & \textbf{7.48}                                           \\ \hline
                    \end{tabular}
                    \captionsetup{belowskip=-20pt}
                    \caption{Results on ScanNet\cite{dai2017scannet}. Grid-GCN achieves $10$$\times$ speed-up on average over other models. Under batch size of 4 and 1, we test our model with $1$$\times$$K$ neighbor nodes. A compact model with $0.5 \times K$ is also reported.}
                    \label{tab:scannet}
                \end{adjustwidth}
            \end{table}
            
            \begin{table}[htb!]
                \begin{adjustwidth}{-5pt}{-5pt}
                \setlength\tabcolsep{3pt}
                \begin{tabular}{l@{\hskip-2pt}ccc@{\hskip1pt}c}
                    \hline
                    \multicolumn{2}{c}{Input (xyzrgb as default)} & mIOU & OA & latency(ms) \\ \hline \hline
                    \multicolumn{5}{c}{mIOU$< 54.0$}                                                      \\
                    PointNet\cite{qi2017pointnet}    & $8 \times 4096$             & 41.09      & -        & 20.9         \\
                    DGCNN\cite{wang2019dynamic}       & $8 \times 4096$             & 47.94      & 83.64    & 178.1        \\
                    SegCloud\cite{tchapmi2017segcloud}      & -                    & 48.92      & -        & -            \\
                    RSNet\cite{huang2018recurrent}       & $8 \times 4096$             & 51.93      & -        & 111.5        \\
                    PointNet++\cite{qi2017pointnet++}        & $8 \times 4096$             & 52.28      & -        &              \\
                    DeepGCNs\cite{li2019deepgcns}        & $1 \times 4096$               & 52.49      & -        & 45.63        \\
                    TanConv\cite{tatarchenko2018tangent}    & $8 \times 4096$              & 52.8       & 85.5     & -            \\
                    \textbf{Grid-GCN}$_{(0.5\times Ch)}$           & $8 \times 4096$             & \textbf{53.21}      & \textbf{85.61}    & \textbf{12.9}        \\ \hline  \hline
                    \multicolumn{5}{c}{mIOU$> 54.0$}                                                      \\
                    3D-UNet\cite{cciccek20163d}       & $8 \times 96^3$ volume               & 54.93      & 86.12    & 574.7        \\
                    PointCNN\cite{li2018pointcnn}       & -                    & 57.26      & 85.91    & -            \\
                    PVCNN++\cite{liu2019point}           & $8 \times 4096$             & 57.63      & 86.87    & 41.1         \\
                    \textbf{Grid-GCN}$_{(1\times Ch)}$               & $8 \times 4096$             & \textbf{57.75}      & \textbf{86.94}    & \textbf{25.9}         \\ \hline
                \end{tabular}
                \captionsetup{aboveskip=5pt}
                \captionsetup{belowskip=-10pt}
                \caption{Results on S3DIS\cite{2017arXiv170201105A} area 5. Grid-GCN is on average $8\times$ faster than other models. We halve the output channels of GridConv for Grid-GCN$_{(0.5\times Ch)}$.  }
                \label{tab:s3dis}
            \end{adjustwidth}
            \end{table}
    \vspace{0pt}
    \subsection{Ablation Studies} \label{ablation}
        \begin{table}[htb!]
            \begin{adjustwidth}{-10pt}{-10pt}
            \setlength\tabcolsep{4pt}
            \begin{tabular}{l@{\hskip1pt}c@{\hskip2pt}c@{\hskip1pt}c@{\hskip1pt}ccc}
                \hline
                & K  & Channels              & Pooling  & Weight & OA   & latency \\ \hline
                Grid-GCN$^0$ & 32 & (32,64,256) & No   &  No          & 91.1 & 15.4ms  \\
                Grid-GCN$^1$ & 32 & (32,64,256) & No   &   Yes          & 91.5 & 15.9ms  \\
                Grid-GCN$^2$ & 32 & (64,128,256) & No  &   Yes          & 92.0 & 21.8ms  \\
                Grid-GCN$^3$ & 64 & (64,128,256) & Yes   &  Yes       & 92.7 & 26.2ms  \\
                Grid-GCN$^{full}$ & 64 & (128,256,512) & Yes & Yes       & 93.1 & 42.2ms  \\ \hline
            \end{tabular}
            \captionsetup{aboveskip=5pt}
            \captionsetup{belowskip=-10pt}
            \caption{Ablation studies on ModelNet40\cite{wu20153d}. Our models have 3 layers of GridConv. K is the number of node points in the first GridConv. We also change the number of the output feature channels from these 3 layers. Grid context pooling (shorted as pooling here) are also removed for Grid-GCN$^{0-2}$. Grid-GCN$^0$ also removes coverage weight in edge relation.}
            \label{tab:ablation}
            \end{adjustwidth}
        \end{table}
        
        In the experiment on ModelNet10 and ModelNet40\cite{wu20153d}, our full model has 3 GridConv layers. As shown in Table \ref{tab:ablation}, we conduct reductions on the number of the output feature channels from GridConv layers, the number of nodes $K$ in the first GridConv layer, and whether to use Grid context pooling and coverage weight. On one hand, reducing the number of channels from Grid-GCN$^{full}$ gives Grid-GCN$^3$ $37\%$ speed-up. On the other hand, reducing $K$ and removing Grid context pooling from Grid-GCN$^3$ doesn't give Grid-GCN$^2$ much speed benefit but incurs a loss on accuracy. This demonstrates the efficiency and effectiveness of CAGQ and Grid context pooling. Coverage weight is useful as well because it introduces little overhead in latency but increases the overall accuracy.
        
        \subsubsection{Scalability Analysis}
        \vspace{-10pt}
            \label{sec:scala}
            \begin{table}[htb!]
                \begin{adjustwidth}{0pt}{-5pt}
                    \setlength\tabcolsep{4pt}
                    \begin{tabular}{llllll}
                        \hline
                        Num. of points ($N$)   & 2048 & 4096 & 16384 & 40960 & 81920 \\
                        Num. of clusters ($M$) & 512  & 1024 & 2048  & 4096  & 8192  \\ \hline
                        PointNet++           & 4.7  & 8.6  & 19.9  & 64.6  & 218.9 \\
                        \textbf{Grid-GCN}             & 4.3  & 4.7  & 8.1   & 12.3  & 19.8  \\ \hline
                    \end{tabular}
                    \captionsetup{aboveskip=5pt}
                    \captionsetup{belowskip=-10pt}
                    \caption{Inference time (ms) on ScanNet\cite{dai2017scannet} under different scales. We compare Grid-GCN with PoinNet++\cite{qi2017pointnet++} on different numbers of input points per scene. The batch size is 1. $M$ is the number of point groups on the first network layer.}
                    \label{tab:scale}
                \end{adjustwidth}
            \end{table}
            We also test our model's scalability by gradually increasing the number of input points on ScanNet \cite{dai2017scannet}. We compare our model with PointNet++ \cite{qi2017pointnet++}, one of the most efficient point-based method. We report the results in Table \ref{tab:scale}. Under the setting of 2048 points, the latency of two models are similar. However, when increasing the input point from 4096 to 81920, Grid-GCN achieves up to $11\times$ speed-up over PointNet++, which shows the dominating capability of our model in processing large-scale point clouds. 
            

\section{Conclusion}
    In this paper, we propose Grid-GCN for fast and scalable point cloud learning. Grid-GCN achieves efficient data structuring and computation by introducing Coverage-Aware Grid Query (CAGQ). CAGQ drastically reduces data structuring cost through voxelization and provides point groups with complete coverage of the whole point cloud. A graph convolution module Grid Context Aggregation (GCA) is also proposed to incorporate the context features and coverage information in the computation. With both modules, Grid-GCN achieves state-of-the-art accuracy and speed on various benchmarks. Grid-GCN, with its superior performance and unparalleled efficiency, can be used in large-scale real-time point cloud processing applications.

{\small
\bibliographystyle{ieee_fullname}
\bibliography{egbib}
}

    \begin{appendices}
    	\section{Time complexity deductions of center sampling/node querying methods}
    	We treat the number of voxel neighbors $\lambda$ as a constant. In addition, the center sampling methods are only used during downsampling GridConv.\\
    	1. The time complexity of center sampling methods:
    	\begin{itemize}[itemsep=-1pt,topsep=0pt,leftmargin=15pt]
    		\item RPS: Random sampling methods such as \cite{ade2012simulated} provide RPS a complexity of $O(min(N, MlogM))$. In practice, $MlogM$ has a same or smaller magnitude of $N$, therefore report $O(N)$ in the time complexity table.
    		\item FPS: FPS on a finite point set has $O(N^2)$, when $M$ is not extremely small. However, \cite{eldar1992irregular} uses a Voronoi diagram to find the area that the point should exist, then find the nearest point in the calculated area. As an approximate algorithm, it has $O(NlogN)$.
    		\item RVS: To sample point groups, CAGQ first scans over all points and build indices, which takes $O(N)$, RVS then randomly samples $M$ center voxels from at most $N$ occupied voxels (the num. of occupied voxels $\leq$ the number of raw points), which takes $O(min(N, MlogM))$. Under the same assumption of RPS, we report $O(N)$ in the time complexity table.
    		\item CAS: If choosing CAS, we need to check all the unpicked occupied voxels and challenge the incumbents. To calculate a pair of $H_{add}$ and $H_{rmv}$, CAGQ checks $\lambda$ voxel neighbors of a challenger and an incumbent, result in a $O(\lambda N) = O(N)$ for all extra operations. Therefore CAS still has a complexity of $O(N)$.
    	\end{itemize} 
    	2. The time complexity of node querying methods:
    	\begin{itemize}[itemsep=-1pt,topsep=0pt,leftmargin=15pt]
    		\item Ball Query: For each center, Ball Query needs to run over $N$ points to collect in-range points, then sample$K$points. Therefore it needs $O(MN)$
    		\item k-NN: For each center, vanilla k-NN picks$K$nearest points from $N$ points. The partition-based topK method takes $O(N)$ computation. Therefore each center has $O(N)$.  k-NN can also first check if a point is within a range, then query top$K$candidate points. These two methods have the same worst-case complexity. The overall complexity is $O(MN)$.
    		\item Cube Query: CAGQ's Cube Query randomly picks$K$points from $\lambda n_v$ context points. Since the order of points in each neighborhood is already randomized during GPU's multithreading collection, the overall complexity is $O(MK)$.
    		\item CAGQ's k-NN: CAGQ picks$K$nearest points among points in the neighborhood. The partition-based topK algorithm provides a $O(\lambda n_v)$ solution for each point group. If $\lambda$ is treated as a constant, the overall complexity is $O(Mn_v)$.
    	\end{itemize}
    	\section{Training Details}
    	For all experiments on ModelNet10, ModelNet40\cite{wu20153d}, ScanNet\cite{dai2017scannet} and S3DIS\cite{2017arXiv170201105A}, we use Adam\cite{kingma2014adam} optimizer with beta1 = 0.9 and beta2 = 0.999. All models use batch normalization\cite{ioffe2015batch} with no  momentum decay and trained on a single RTX 2080 GPU.
    	
    	For ModelNet10 and ModelNet40\cite{wu20153d}, we start with a learning rate of 0.001 and reduce the learning rate by a factor of 0.7 every 60 epochs and stop at 330 epochs. We don't apply weight decay. The network has 2 downsampling GridConv layers each has 1024 and 128 point groups and a final global GridConv layer to group all points as one graph. 
    	
    	For ScanNet\cite{dai2017scannet}, we start with a learning rate of 0.001 and reduce the learning rate by a factor of 0.7 every 150 epochs and stop at 1500 epochs. Please note during each epoch, we only sample one block on the fly in each region. Therefore the training of one epoch is very quick. The network has 3 downsampling GridConv layers each has 1024, 256 and 24 point groups and 3 upsampling GridConv layers. We set our weight decay as $10^{-5}$ during training. 
    	
    	For S3DIS\cite{2017arXiv170201105A}, we start with a learning rate of 0.001 and reduce the learning rate by a factor of 0.8 every 40 epochs and stop at 200 epochs. The network has 3 downsampling GridConv layers each has 512, 256 and 24 point groups and 3 upsampling GridConv layers. We set our weight decay as $10^{-8}$ during training. 
    	\section{Comparison between CAGQ and naive Grid Query}
    	In 2D image learning, a convolution kernel usually traverses with a stride size that is smaller than the kernel size, leading to overlapping receptive fields. Since naive Grid Query first voxelizes the space and randomly picks $M$ voxels, and then samples $K$ points only within the voxel, the queried point groups have no space overlaps. On the other hand, CAGQ queries points inside voxel neighbors and utilizes Coverage-aware sampling to make the center voxels more evenly distributed.
    	
    	To show the advantage of CAGQ over naive Grid Query, we compare 3 models on ScanNet\cite{dai2017scannet} and report the results in Table \ref{tb:gq}. The two models using CAGQ are also the models we report in section 5.2 of the paper. We also train a model using naive Grid Query. As a result of the non-overlapping coverage by its point groups, the overall accuracy of the model with naive Grid Query can hardly reach $80\%$. 
    	\begin{table}[h!]
    		\begin{tabular}{lcc}
    			\hline
    			& OA    & Latency (ms) \\ \hline
    			Grid-GCN(Grid Query + $1\times K$) & 79.9$\%$ & 15.9         \\
    			Grid-GCN(CAGQ + $0.5\times K$)           & 83.9$\%$ & 16.6         \\
    			Grid-GCN(CAGQ + $1\times K$)             & 85.4$\%$ & 20.8         \\ \hline
    		\end{tabular}
    		\caption{The overall accuracy and latency of three Grid-GCN models on ScanNet\cite{dai2017scannet}. Our full model uses CAGQ with $1$$\times$$K$ node points in each group. A compact model with $0.5 \times K$ is also reported. Another model uses naive Grid Query with $1$$\times$$K$ node points.}
    		\label{tb:gq}
    	\end{table}
    	\section{Algorithms of CAGQ}
    	The general procedure of CAGQ is listed as Algorithm \ref{algogen}. The CAGQ k-NN algorithm mentioned in the paper is listed as Algorithm \ref{algoknn}. To use RVS or CAS, we can embed the chosen algorithm in to step 2 of Algorithm \ref{algogen}. Cube Query and k-NN can be embedded in to step 3. 
    	
    	The k-NN methods in Algorithm \ref{algoknn} is efficient in three aspects:
    	\begin{itemize}[itemsep=-1pt,topsep=0pt,leftmargin=15pt]
    		\item instead of all points, the candidates of k-NN are only the points in the neighborhood.  
    		\item We collect $K$ nearest neighbor points first from inner voxel layers, then the outer voxel layers. We can stop at a layer if we have already got $K$ points, since all the points in outer voxel layers are farther away than the point collected so far.
    		\item In each layer of voxel neighbors, if the number of points so far collected plus the points in this layer is less than $K$, we do not need to sort them but can directly include all points in this layer. 
    	\end{itemize}
    	\begin{algorithm}
    		\SetAlgoLined
    		\hspace{-8pt}\textbf{Input} $N$ points: $p_i(\chi_i,w_i), i \in (1,...,N)$ \\
    		\hspace{-8pt}\textbf{Parameters} $N$, $M$, $K$, $\lambda$, $n_v$\\
    		\hspace{-8pt}\textbf{1} Build voxel-point index $Vid$, collect $O_v$:  \\
    		\hspace{0pt}\textbf{For each}  $p_i$ : \\
    		\hspace{12pt}$(u,v,w)\leftarrow$ quantize $p_i(x,y,z)$ into voxel index \\
    		\hspace{12pt}\textbf{If} voxel$(u,v,w)$ is first visited \\
    		\hspace{27pt}         Add $(u,v,w)$ into $O_v$ \\    
    		\hspace{13pt}\textbf{If} $Vid_{(u,v,w)}$ stores fewer than $n_v$ points \\
    		\hspace{27pt}Push point index into $Vid_{(u,v,w)}$ \\
    		\hspace{-8pt}\textbf{2} Center voxel sampling: \\
    		\hspace{0pt} $O_c\leftarrow$ select $M$ voxels from $O_v$, by RVS or CAS. \\
    		\hspace{-8pt}\textbf{3} Query node points and calculate group centers: \\
    		\hspace{0pt} \textbf{For each} center voxel $v_j$ in $O_c$ \\
    		\hspace{12pt}Retrieve points in $\pi(v_j)$ by using indices.\\
    		\hspace{12pt}Pick node points $\{p_{j1},p_{j2},..p_{jK}\}$ in the neighborhood by using Cube Query or k-NN \\
    		\hspace{12pt}$w_{c_j} \leftarrow \sum_{k=1}^{K} w_{jk} $ \\
    		\hspace{12pt}$\chi_{c_j} \leftarrow$ weighted mean of $\{\chi_{j1},...,\chi_{jK}\}$ \\
    		\hspace{-8pt}\textbf{Return} $M$ point groups: $group_j:$ $( c_j(\chi_{c_j},w_{c_j}), \{p_{j1},p_{j2},..p_{jK}\} ), j \in (1,...,M)$
    		\caption{CAGQ general procedure}
    		\label{algogen}
    	\end{algorithm}
    	\begin{algorithm}
    		\SetAlgoLined
    		\hspace{-8pt}\textbf{Input} A center voxel $v_j$, voxel-point index $Vid, O_v$ \\
    		\hspace{-8pt}\textbf{Parameters} $K$ \\
    		\hspace{-8pt}Counter = 0; node points =$\{\}$  \\
    		\hspace{-8pt}\textbf{For each} level$_i$ of $\pi(v_j)$ (level$_0$ is the center voxel $v_j$ itself, level$_1$ is the surrounding 26 voxels, etc.): \\
    		\hspace{0pt} $kl = 0$ \\
    		\hspace{0pt}LayerPoints = $\{\}$ \\
    		\hspace{0pt}\textbf{For each} voxel $v_{l}$ in level$_i$: \\
    		\hspace{12pt}\textbf{If} $v_{l} \in O_v$:\\
    		\hspace{27pt}stored points $\leftarrow$ Retrieve points from $Vid(v_{l})$:\\
    		\hspace{27pt}LayerPoints $\leftarrow$ add stored points \\
    		\hspace{27pt}$kl += |$stored points$|$ \\
    		\hspace{0pt}$topkl = min(K$-Counter, $kl)$ \\
    		\hspace{0pt}\textbf{If} $topkl = kl$: \\ 
    		\hspace{12pt}node points $\leftarrow$ LayerPoints \\
    		\hspace{0pt}\textbf{Else}: \\ 
    		\hspace{12pt}node points $\leftarrow$ topK(LayerPoints, $topkl$) \\
    		\hspace{0pt}Counter += $topkl$ \\
    		\hspace{0pt}\textbf{If} Counter $\geq$ $K$: \\
    		\hspace{12pt}break; \\
    		\hspace{-8pt}\textbf{Return} node points
    		\caption{CAGQ k-NN for one point group in step 3 of Algorithm \ref{algogen}}
    		\label{algoknn}
    	\end{algorithm} 
    	\onecolumn
    	\section{Calculation of center}
    	In a point group, we calculate $w_c$ as the sum of its node points' coverage weight. $\chi_c$ is calculated as the barycenter of these nodes, weighted by the coverage weight.  
    	\begin{adjustwidth}{-0pt}{0pt}
    		\setlength{\abovedisplayskip}{0pt}%
    		\setlength{\abovedisplayshortskip}{\abovedisplayskip}%
    		\setlength{\belowdisplayskip}{5pt}%
    		\begin{align}
    		w_c &= \sum_{j=1}^{K} w_j\\
    		\chi_c(x,y,z) &= \cfrac{\sum_{j=1}^{K} w_j \cdot \chi_j(x,y,z)}{\sum_{j=1}^{K} w_j} 
    		\end{align}
    	\end{adjustwidth}
    	\section{Performance comparisons of data structuring methods (more conditions)}
    	We list the full experiments of different data structuring methods' coverage and latency under more conditions in Table \ref{tb:morecond}. The first section shows the coverage of occupied voxels. We only report the space coverage of center sampling methods+Ball Query or Cube Query, because the purpose of k-NN's is not to query node points that is evenly spread, but to query the nearest neighbor node points. The second and third section report the latency. 
    	\begin{table*}[htb!]
    		\setlength\tabcolsep{5pt}
    		\begin{tabular}{ccc|cccc|cccc|cccc}
    			\hline
    			\multicolumn{3}{c|}{Center sampling}   & RPS  & FPS           & RVS*         & CAS*          & RPS           & FPS    & RVS*          & CAS*  & RPS    & FPS    & RVS*           & CAS*  \\ \hline
    			\multicolumn{3}{c|}{Neighbor querying} & Ball & Ball          & Cube         & Cube          & Ball          & Ball   & Cube          & Cube  & k-NN   & k-NN   & k-NN           & k-NN  \\ \hline
    			N                       &$K$  &$M$    & \multicolumn{4}{c|}{Occupied space coverage($\%$)}     & \multicolumn{8}{c}{Latency (ms) with batch size = 1}                                     \\ \hline
    			\multirow{9}{*}{1024}   & 8   & 8     & 12.3 & 12.9          & 13.1         & \textbf{14.9} & \textbf{0.29} & 0.50    & 0.51          & 0.64  & 0.84   & 0.85   & \textbf{0.51}  & 0.65  \\
    			& 32  & 8     & 22.9 & 21.4          & 22.4         & \textbf{31.7} & \textbf{0.34} & 0.51   & 0.51          & 0.69  & 2.12   & 1.96   & \textbf{0.63}  & 0.71  \\
    			& 128 & 8     & 22.3 & 22.6          & 23.5         & \textbf{34.2} & \textbf{0.34} & 0.51   & 0.94          & 1.04  & 8.26   & 6.70    & \textbf{1.41}  & 1.63  \\
    			& 8   & 32    & 34.4 & 43.7          & 40.0           & \textbf{45.6} & \textbf{0.31} & 0.53   & 0.51          & 0.65  & 1.31   & 1.36   & \textbf{0.57}  & 0.69  \\
    			& 32  & 32    & 58.2 & 69.48         & 60.1         & \textbf{73.0}   & \textbf{0.36} & 0.55   & 0.53          & 0.57  & 4.68   & 4.72   & \textbf{0.93}  & 0.68  \\
    			& 128 & 32    & 60.0   & 70.1          & 61.3           & \textbf{74.7} & \textbf{0.37} & 0.53   & 0.96          & 1.08  & 22.23  & 21.08  & \textbf{2.24}  & 2.58  \\
    			& 8   & 128   & 64.0   & 72.5          & 82.3         & \textbf{85.6} & \textbf{0.32} & 0.78   & 0.44          & 0.58  & 1.47   & 1.74   & \textbf{0.52}  & 0.61   \\
    			& 32  & 128   & 92.7 & 98.9          & 95.3         & \textbf{99.6} & \textbf{0.38} & 0.81   & 0.50          & 0.62  & 5.34   & 5.66   & \textbf{1.18}  & 1.39  \\
    			& 128 & 128   & 93.6 & 99.5          & 95.8         & \textbf{99.7} & \textbf{0.38} & 0.69   & 0.97          & 0.97  & 32.48  & 32.54  & \textbf{6.85}  & 6.94  \\ \hline
    			\multirow{9}{*}{8192}   & 8   & 64    & 19.2 & 22.9          & 22.1         & \textbf{25.1} & \textbf{0.64} & 1.16   & 0.66          & 0.82  & 1.58   & 1.80    & \textbf{0.65}  & 0.76  \\
    			& 32  & 64    & 42.7 & 42.7          & 35.8         & \textbf{46.3} & 1.47          & 1.47   & \textbf{1.15} & 1.39  & 2.73   & 2.73   & \textbf{1.72}  & 2.16  \\
    			& 128 & 64    & 40.6 & 47.1          & 38.6         & \textbf{51.3} & \textbf{1.14} & 1.55   & 1.18          & 1.38  & 13.70   & 12.72  & \textbf{9.42}  & 11.71 \\
    			& 8   & 256   & 60.1 & 64.1          & 73.3         & \textbf{94.3} & \textbf{0.40}  & 1.51   & 0.53          & 0.61  & 4.53   & 5.54   & \textbf{0.54}  & 0.68  \\
    			& 32  & 256   & 75.4 & 88.4          & 77.6         & \textbf{90.7} & 1.11          & 2.19   & \textbf{1.04} & 1.29  & 5.13   & 5.94   & \textbf{3.06}  & 3.52  \\
    			& 128 & 256   & 79.9 & 90.7          & 80.0           & \textbf{93.5} & 1.19          & 1.19   & \textbf{1.17} & 1.31  & 21.5   & 21.5   & \textbf{15.19} & 17.38 \\
    			& 8   & 1024  & 82.9 & \textbf{96.8} & 92.4         & 94.4          & 0.81          & 4.90    & \textbf{0.54} & 0.77  & 1.53   & 5.36   & \textbf{0.93}  & 0.97  \\
    			& 32  & 1024  & 96.3 & 97.8          & 99.3         & \textbf{99.9} & 1.15          & 5.09   & \textbf{1.10}  & 1.54  & 5.18   & 8.99   & \textbf{4.92}  & 6.32  \\
    			& 128 & 1024  & 98.8 & 99.9          & 99.5         & \textbf{100.0}  & \textbf{1.22} & 5.25   & 1.40          & 1.76  & 111.42  & 111.74  & \textbf{24.18} & 26.45 \\ \hline
    			\multirow{9}{*}{81920}  & 8   & 256   & 21.7 & 25.7          & 26.2         & \textbf{31.2} & 3.46          & 10.54  & \textbf{2.20} & 2.87  & 9.77   & 15.97  & \textbf{1.85}  & 2.15  \\
    			& 32  & 256   & 34.2 & 40.1          & 36.0           & \textbf{48.5} & 7.59          & 14.51  & \textbf{3.15} & 4.34  & 20.43  & 26.14  & \textbf{5.95}  & 6.17  \\
    			& 128 & 256   & 36.6 & 42.6          & 37.4         & \textbf{51.1} & 9.41          & 15.91  & \textbf{3.52} & 4.19  & 77.68  & 78.34  & \textbf{34.04} & 40.04 \\
    			& 8   & 1024  & 50.7 & 63.8          & 67.4         & \textbf{76.0}   & 4.73          & 30.79  & \textbf{2.13} & 2.34  & 10.01  & 35.18  & \textbf{1.84}  & 2.02  \\
    			& 32  & 1024  & 70.6 & 86.3          & 78.3         & \textbf{91.6} & 8.30          & 33.52  & \textbf{3.34} & 3.88  & 19.49  & 43.69  & \textbf{8.76}  & 10.05 \\
    			& 128 & 1024  & 72.7 & 88.2          & 79.1         & \textbf{92.6} & 9.68          & 34.72  & \textbf{4.32} & 4.71  & 71.99  & 93.02  & \textbf{50.70}  & 51.94 \\
    			& 8   & 10240 & 98.8 & 99.2          & \textbf{100.0} & \textbf{100.0}  & 8.82          & 255.9  & \textbf{4.11} & 8.23  & 19.96  & 268.22 & \textbf{9.54}  & 14.88 \\
    			& 32  & 10240 & 98.8 & 99.2          & \textbf{100.0} & \textbf{100.0}  & 8.93          & 260.48 & \textbf{4.22} & 9.35  & 20.38  & 272.48 & \textbf{9.65}  & 17.44 \\
    			& 128 & 10240 & 99.7 & \textbf{100.0}  & \textbf{100.0} & \textbf{100.0}  & 10.73         & 258.49 & \textbf{5.83} & 11.72 & 234.19 & 442.87 & \textbf{69.02} & 83.32 \\ \hline
    		\end{tabular}
    		\caption{Performance comparisons of data structuring methods, run on ModelNet40 \cite{wu20153d}. Center sampling methods include
    			RPS, FPS, CAGQ's RVS and CAS. Neighbor querying methods include Ball Query, Cube Query and k-Nearest Neighbors.
    			Condition variables include $N$ points, $M$ groups, and $K$ node points per group. Occupied space coverage = num. of occupied
    			voxels of queried points / num. of occupied voxels of the original $N$ points.}
    		\label{tb:morecond}
    	\end{table*}
    	\pagebreak
    	\section{Performance on each object class and more visual results of S3DIS}
\begin{figure*}[b!]
    		\setlength{\abovedisplayskip}{0pt}%
    		\setlength{\abovedisplayshortskip}{\abovedisplayskip}%
    		\setlength{\belowdisplayskip}{5pt}%
    		\begin{adjustwidth}{-5pt}{-2pt}
    			\centering
    			\begin{subfigure}{0.33\linewidth}
    				\centering
    				\includegraphics[width=1\linewidth]{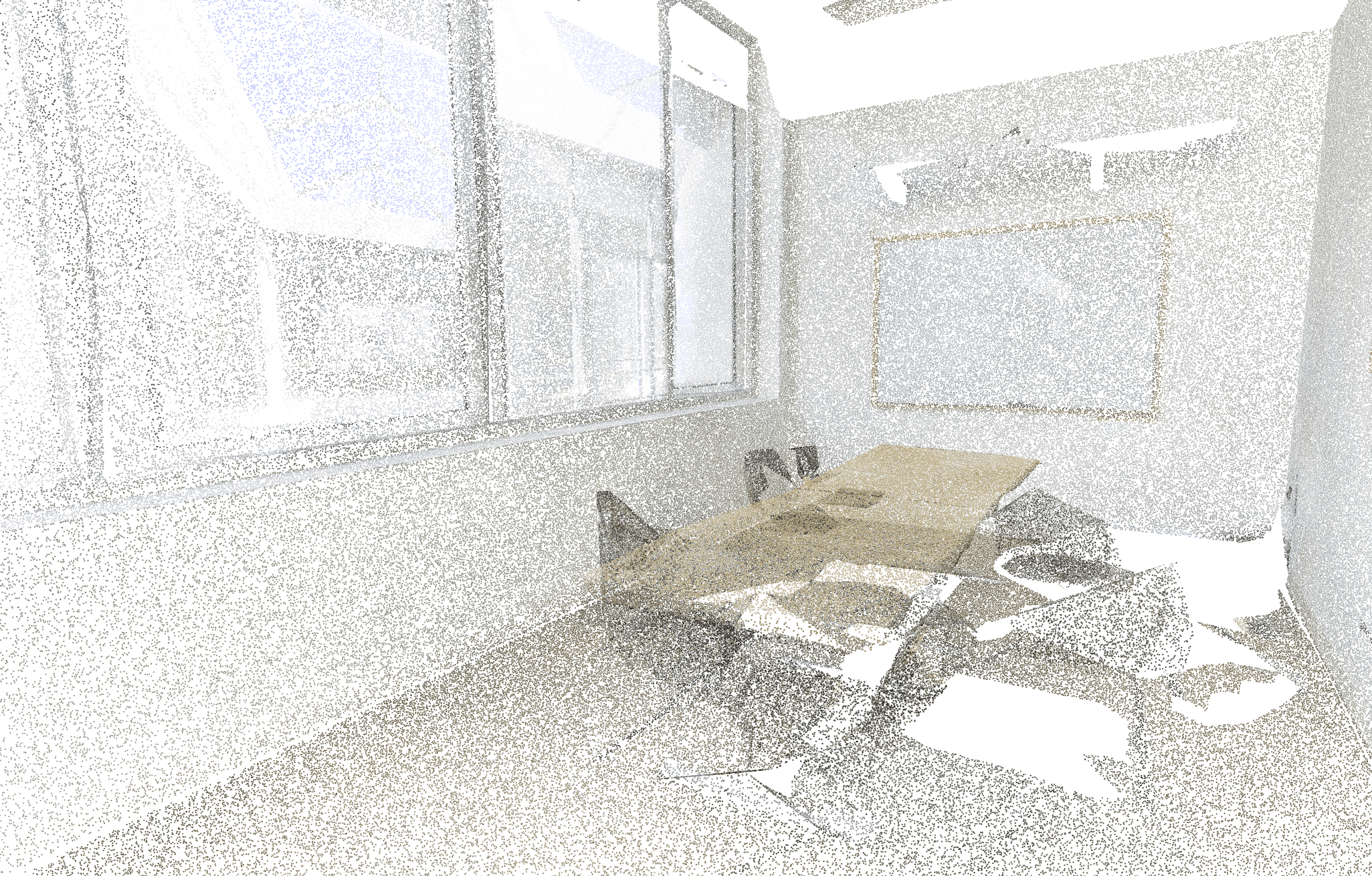}
    			\end{subfigure} \hfill
    			\begin{subfigure}{0.33\linewidth}
    				\centering
    				\includegraphics[width=1\linewidth]{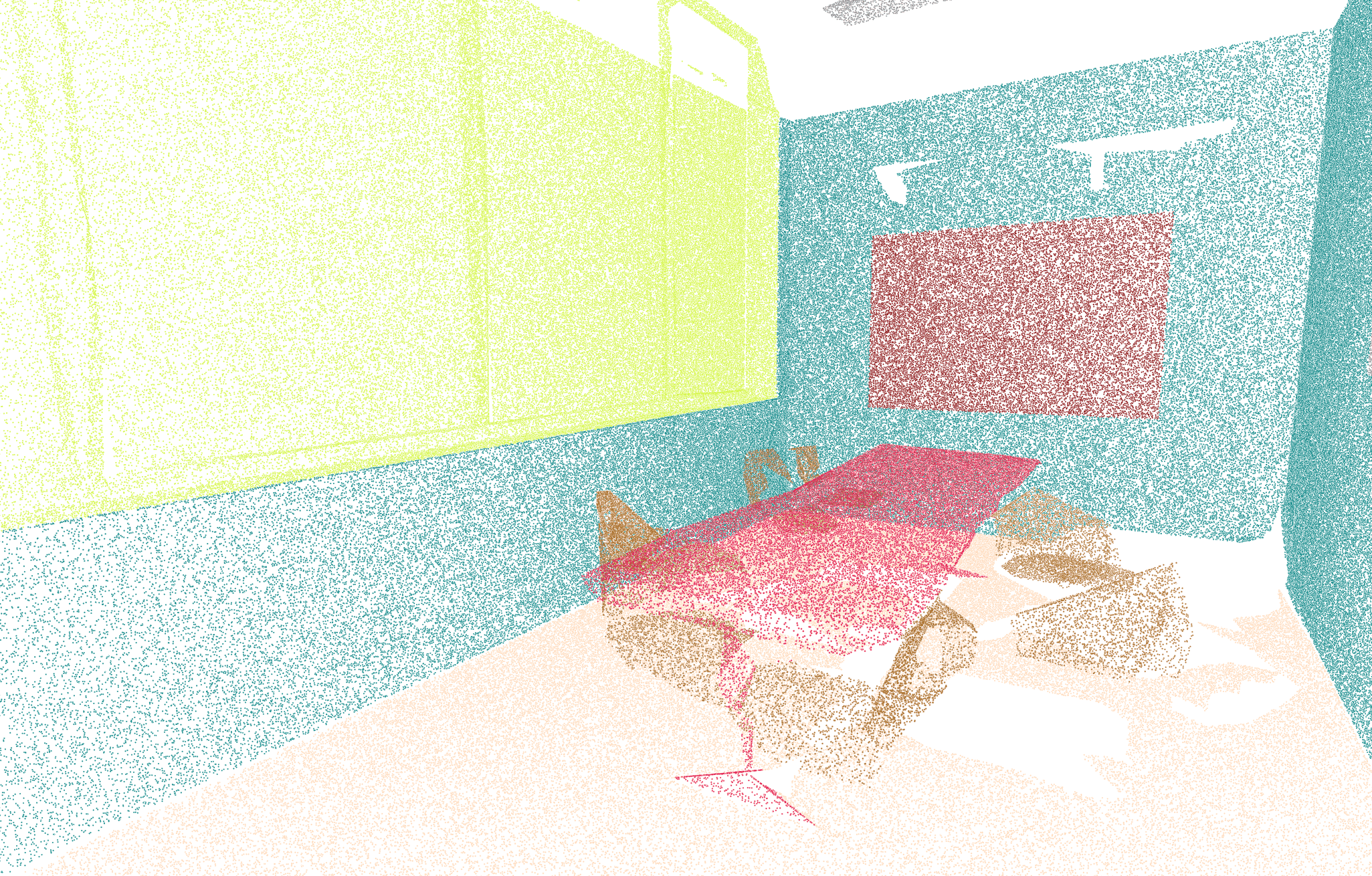}
    			\end{subfigure}
    			\begin{subfigure}{0.33\linewidth}
    				\centering
    				\includegraphics[width=1\linewidth]{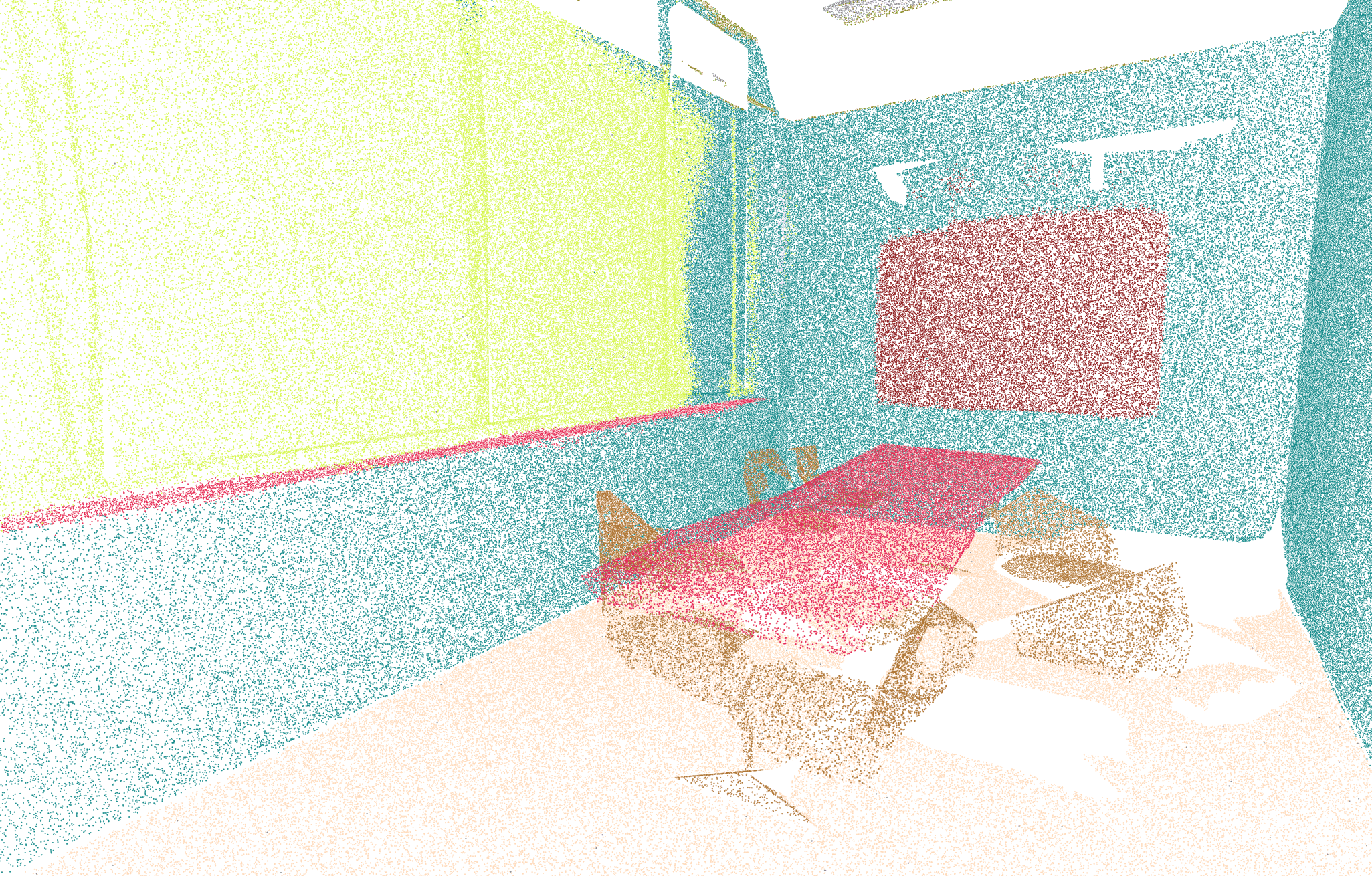}
    			\end{subfigure} \\
    			
    			\begin{subfigure}{0.33\linewidth}
    				\centering
    				\includegraphics[width=1\linewidth]{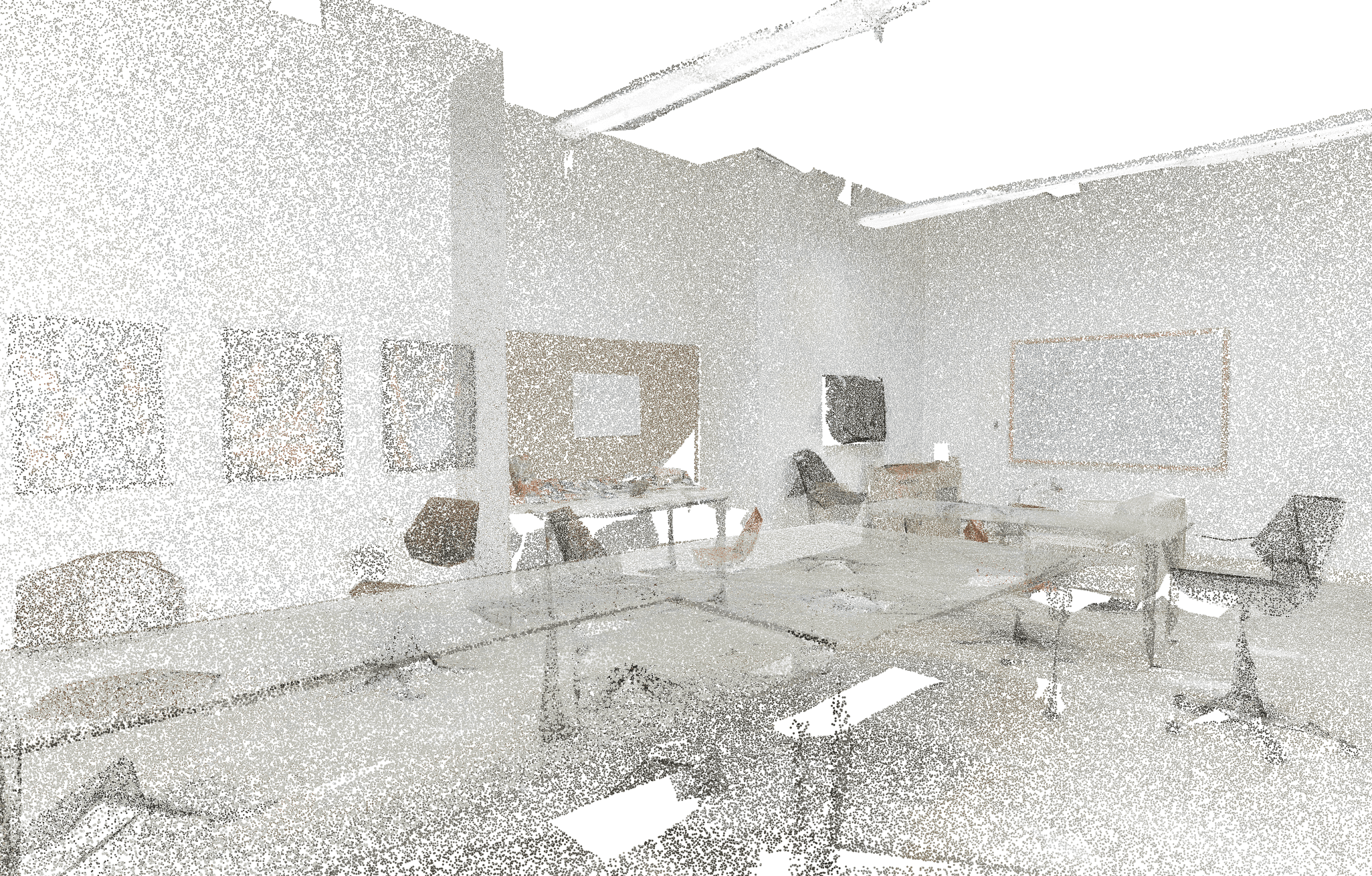}
    			\end{subfigure} \hfill
    			\begin{subfigure}{0.33\linewidth}
    				\centering
    				\includegraphics[width=1\linewidth]{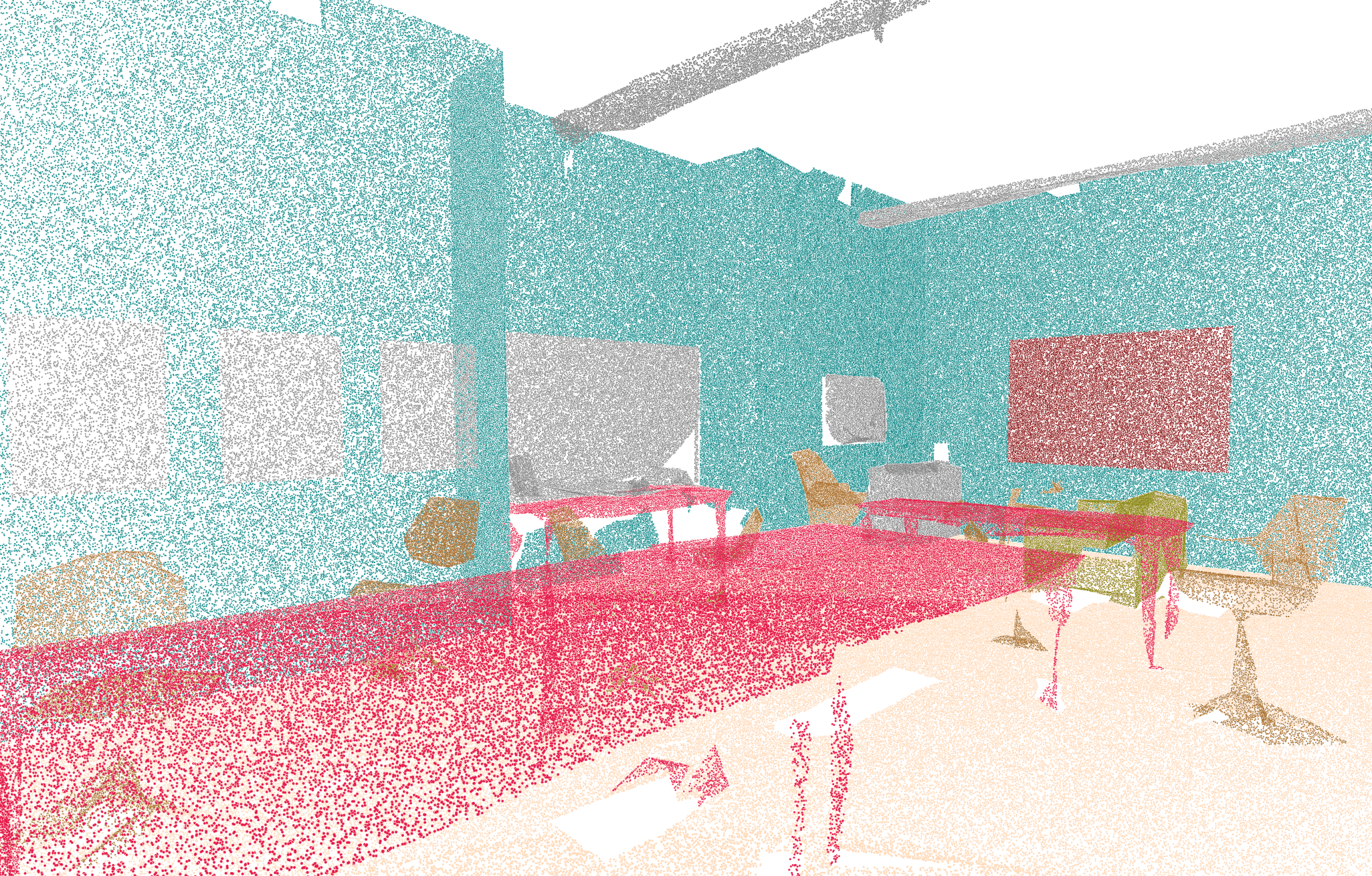}
    			\end{subfigure}
    			\begin{subfigure}{0.33\linewidth}
    				\centering
    				\includegraphics[width=1\linewidth]{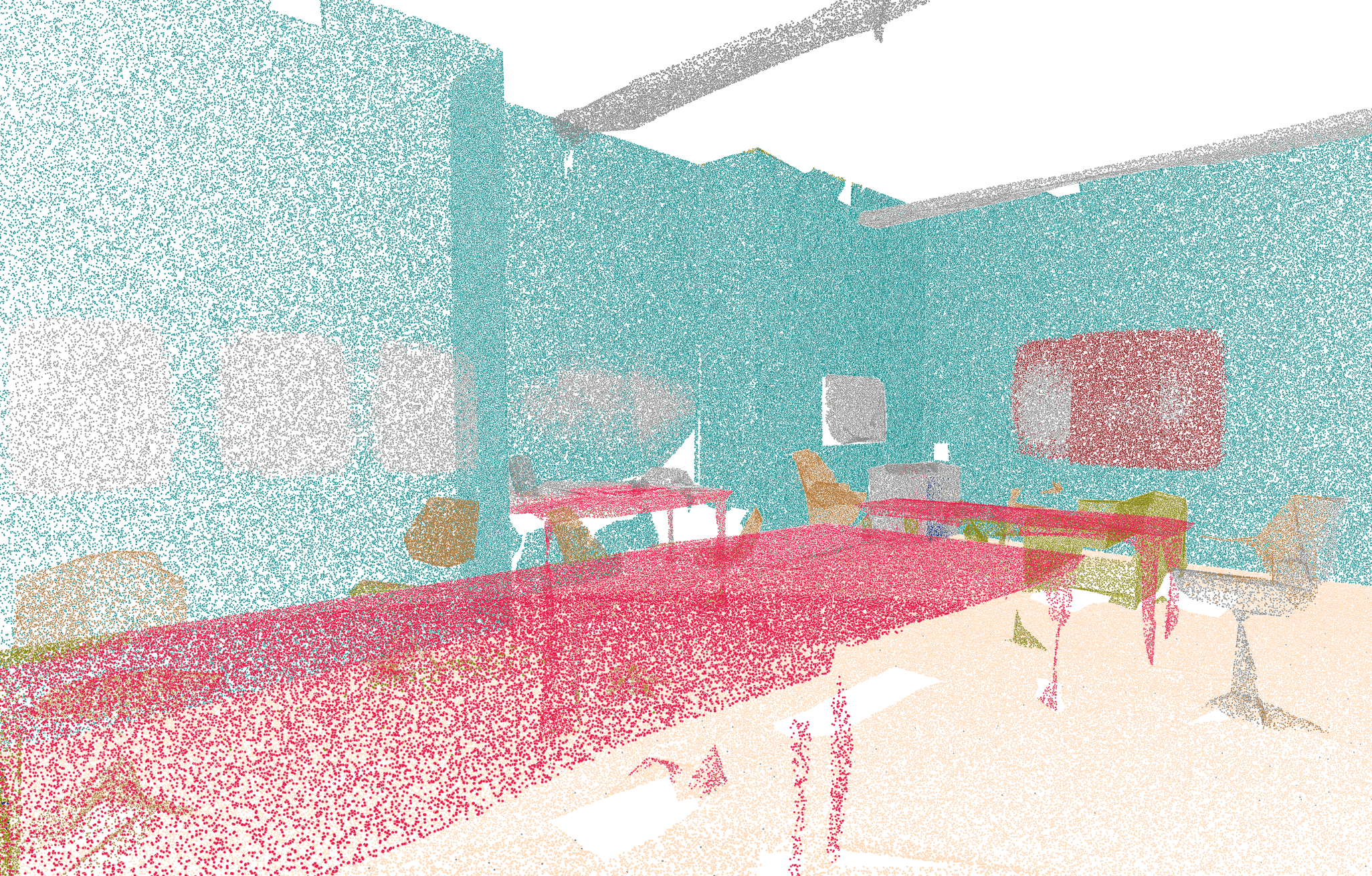}
    			\end{subfigure} \\
    			
    			\begin{subfigure}{0.33\linewidth}
    				\centering
    				\includegraphics[width=1\linewidth]{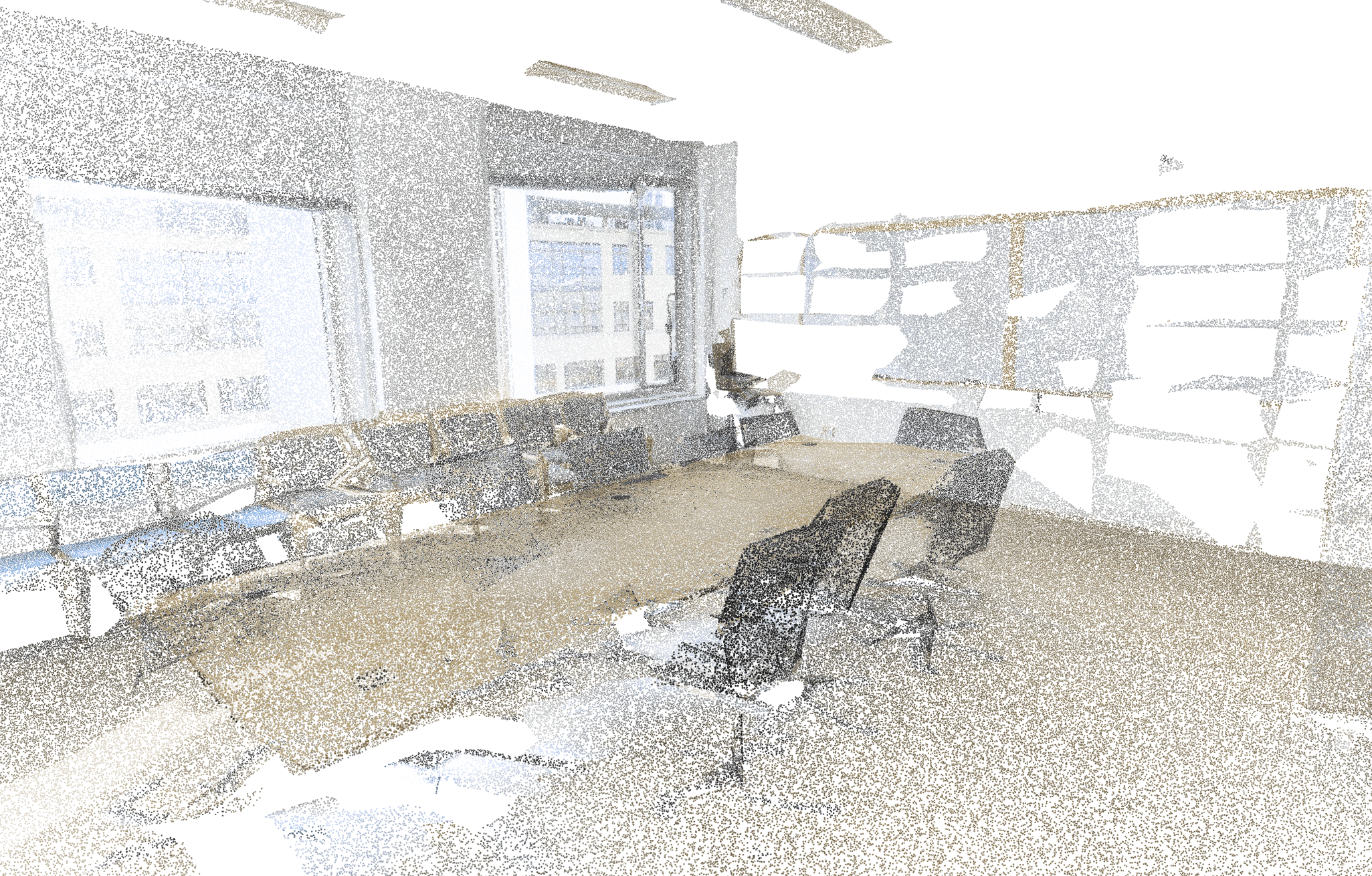}
    			\end{subfigure} \hfill
    			\begin{subfigure}{0.33\linewidth}
    				\centering
    				\includegraphics[width=1\linewidth]{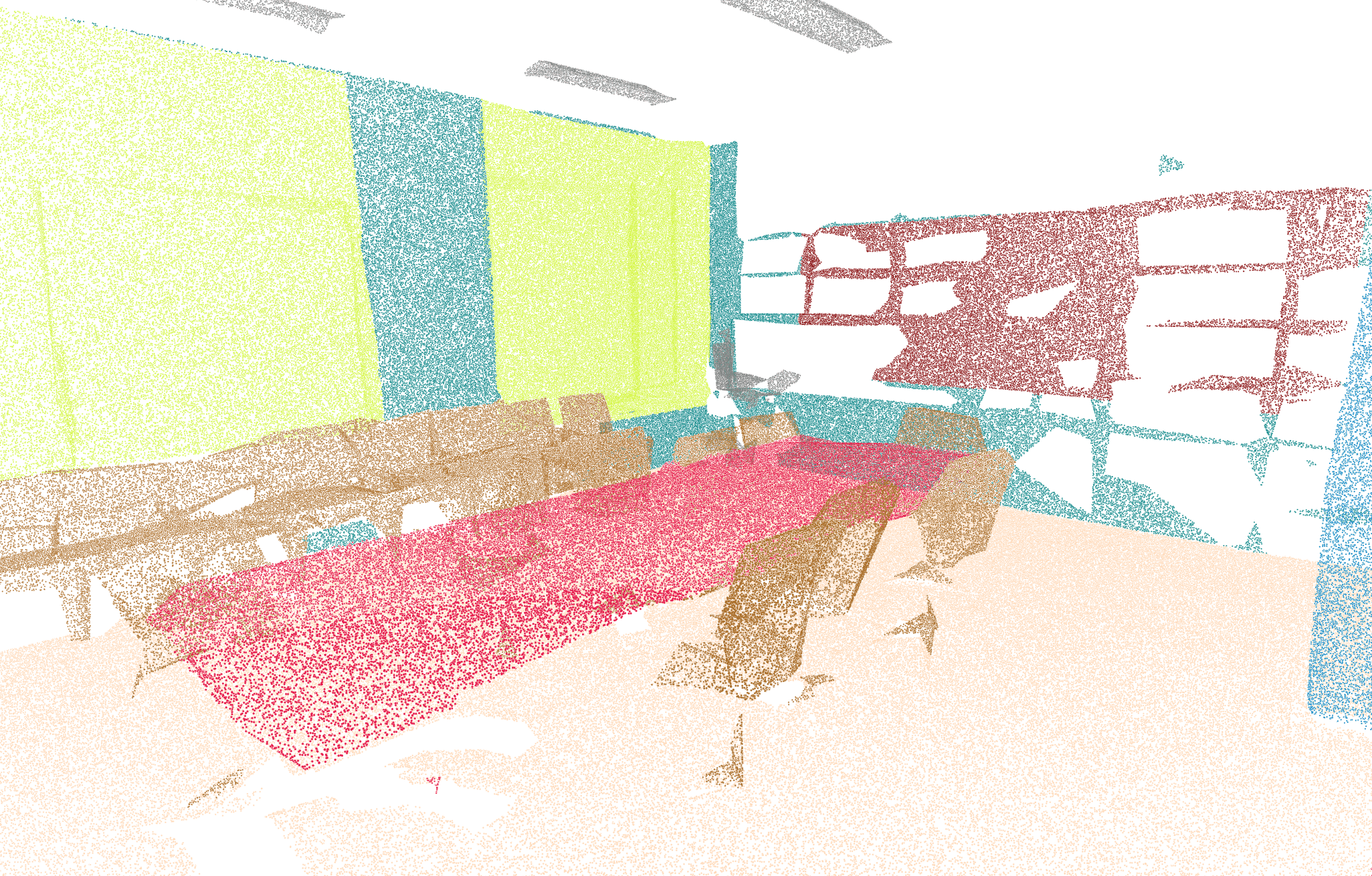}
    			\end{subfigure}
    			\begin{subfigure}{0.33\linewidth}
    				\centering
    				\includegraphics[width=1\linewidth]{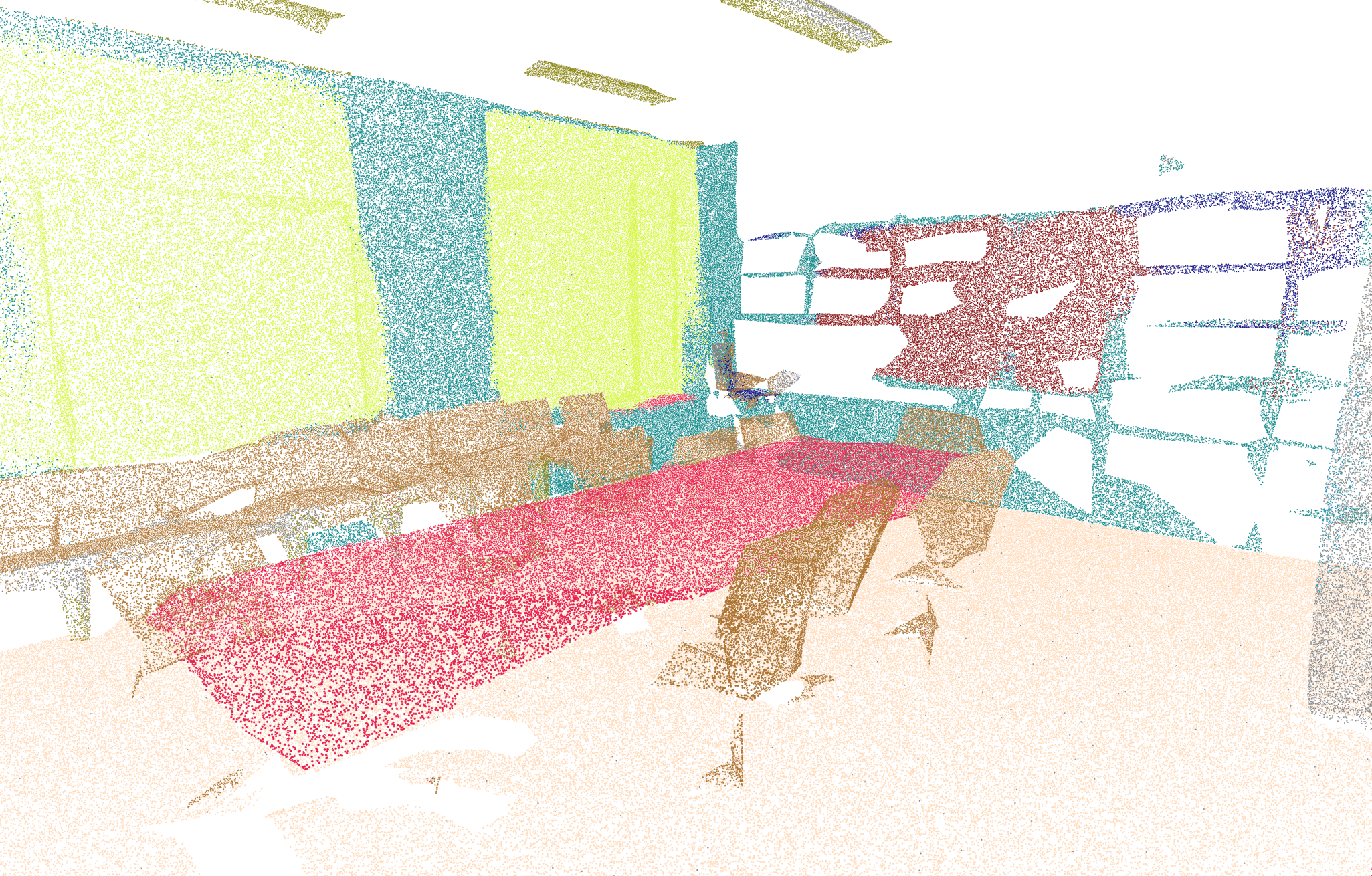}
    			\end{subfigure} \\
    			
    			\begin{subfigure}{0.33\linewidth}
    				\centering
    				\includegraphics[width=1\linewidth]{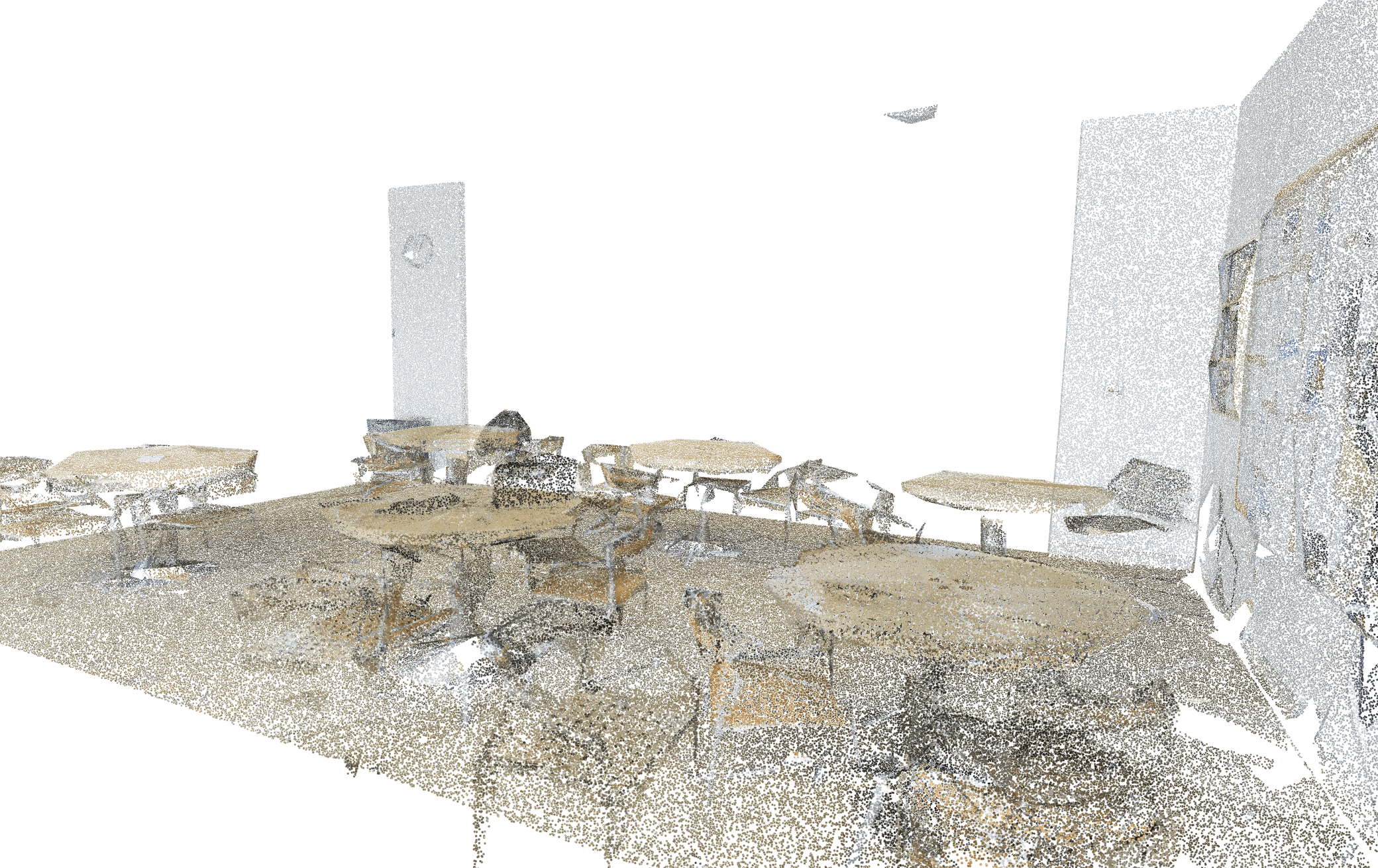}
    				\caption{Input XYZ+RGB}
    			\end{subfigure} \hfill
    			\begin{subfigure}{0.33\linewidth}
    				\centering
    				\includegraphics[width=1\linewidth]{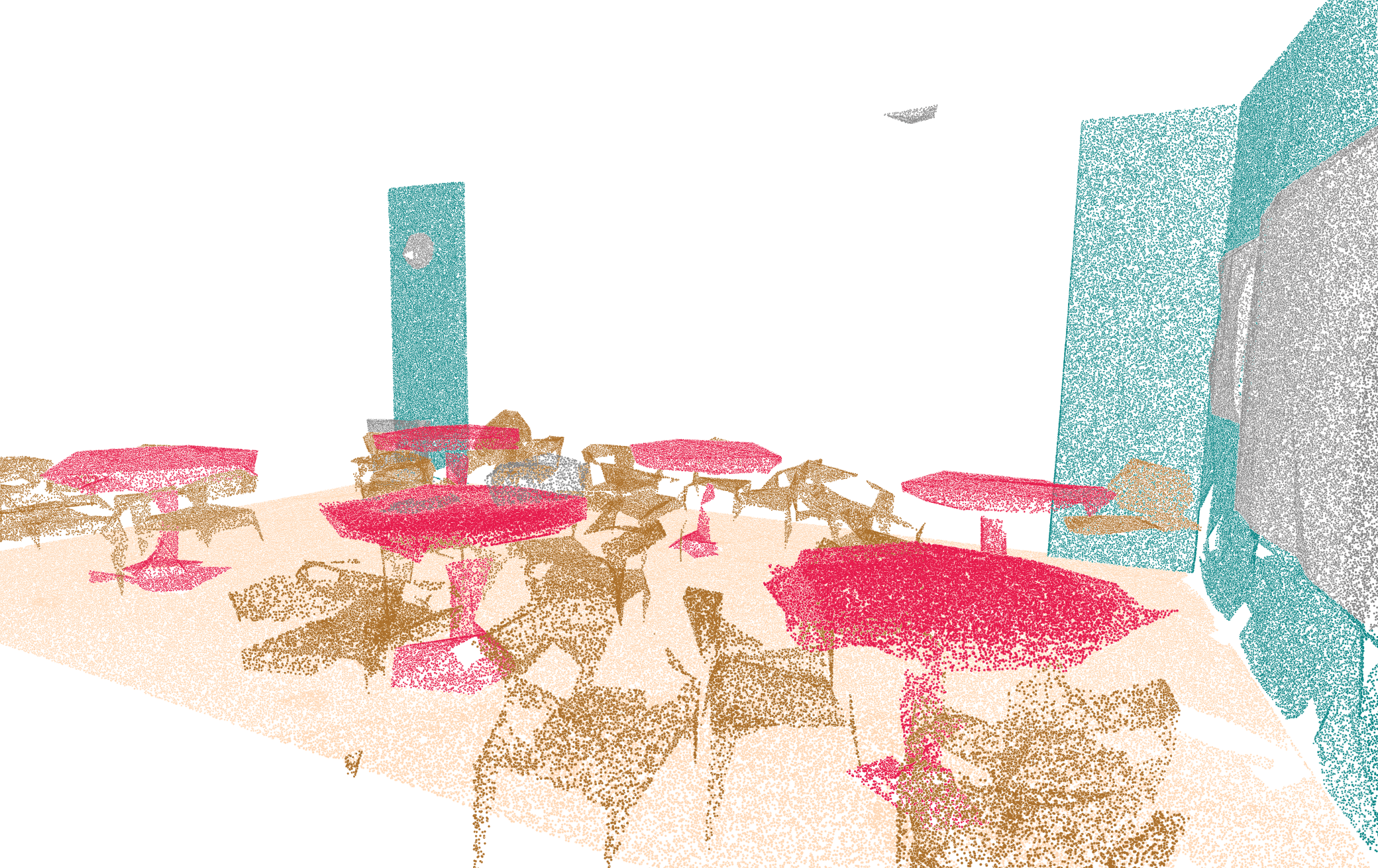}
    				\caption{Ground truth}
    			\end{subfigure}
    			\begin{subfigure}{0.33\linewidth}
    				\centering
    				\includegraphics[width=1\linewidth]{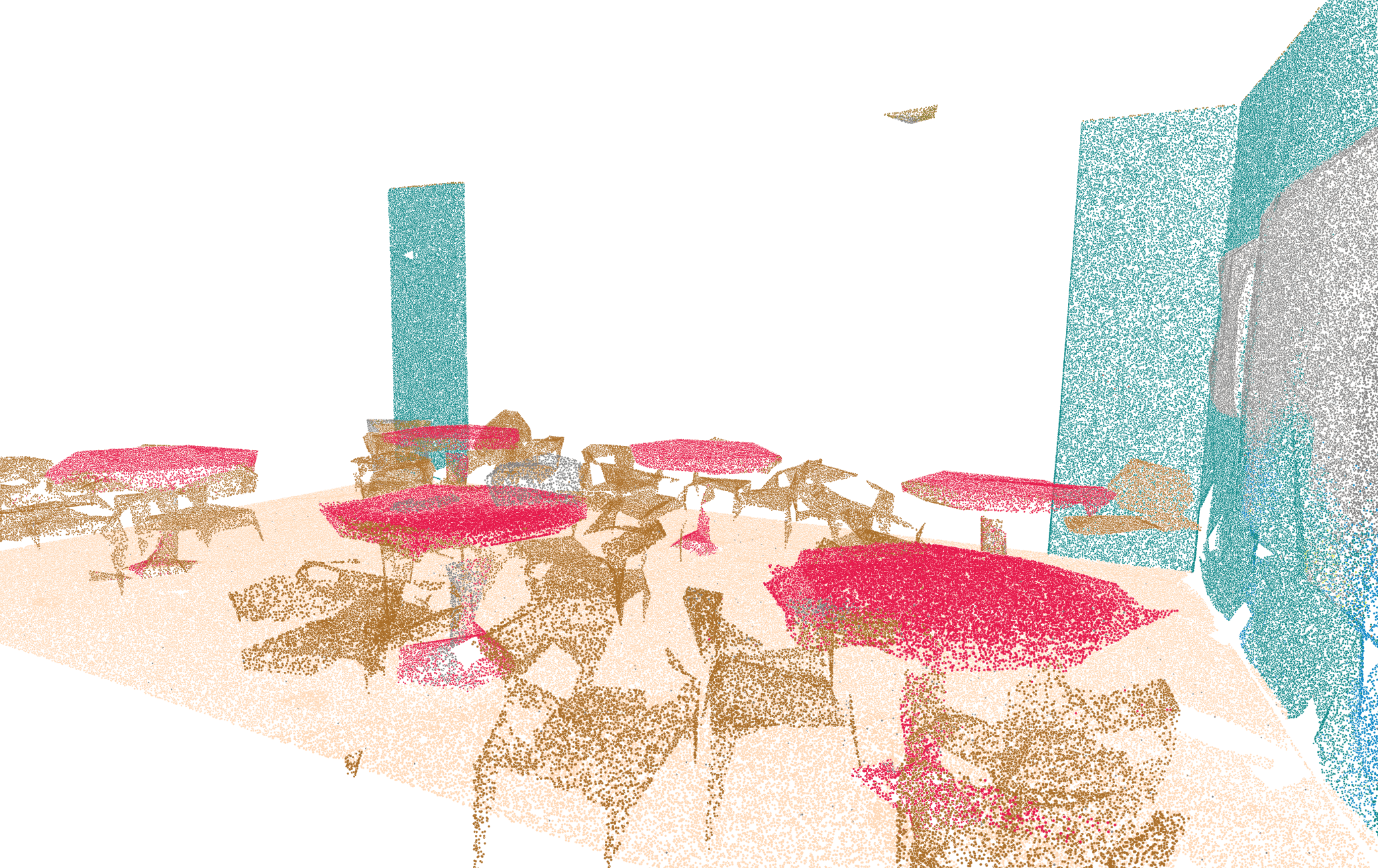}
    				\caption{Ours}
    			\end{subfigure}
    			\captionsetup{aboveskip=5pt}
    			\captionsetup{belowskip=0pt}
    		\end{adjustwidth}
    		\caption{More visual result of S3DIS\cite{2017arXiv170201105A} area 5}
    		\label{fig:more1}
    	\end{figure*}

\begin{figure*}[hbt!]
    		\setlength{\abovedisplayskip}{0pt}%
    		\setlength{\abovedisplayshortskip}{\abovedisplayskip}%
    		\setlength{\belowdisplayskip}{5pt}%
    		\begin{adjustwidth}{-5pt}{-2pt}
    			\centering
    			\begin{subfigure}{0.33\linewidth}
    				\centering
    				\includegraphics[width=1\linewidth]{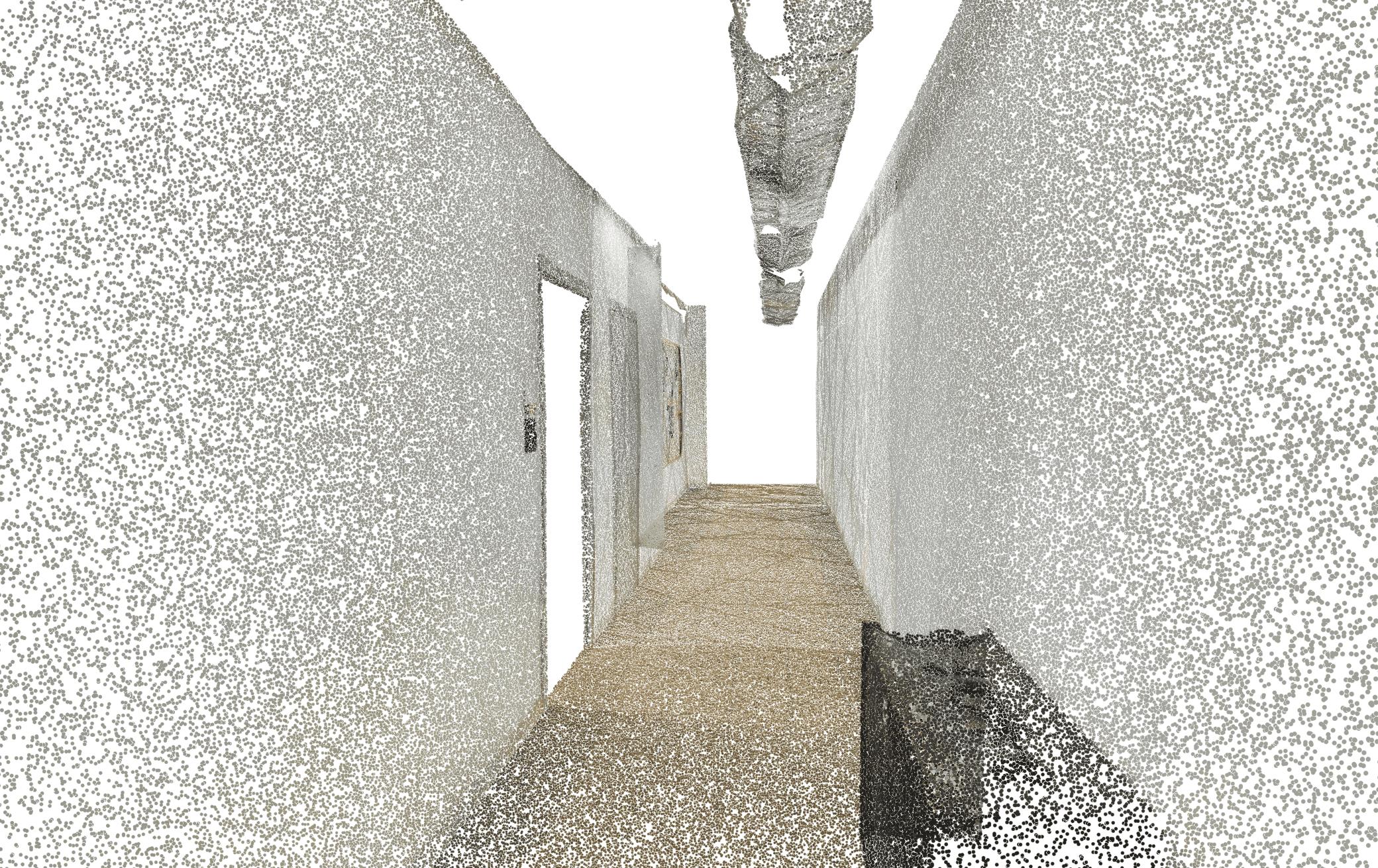}
    			\end{subfigure} \hfill
    			\begin{subfigure}{0.33\linewidth}
    				\centering
    				\includegraphics[width=1\linewidth]{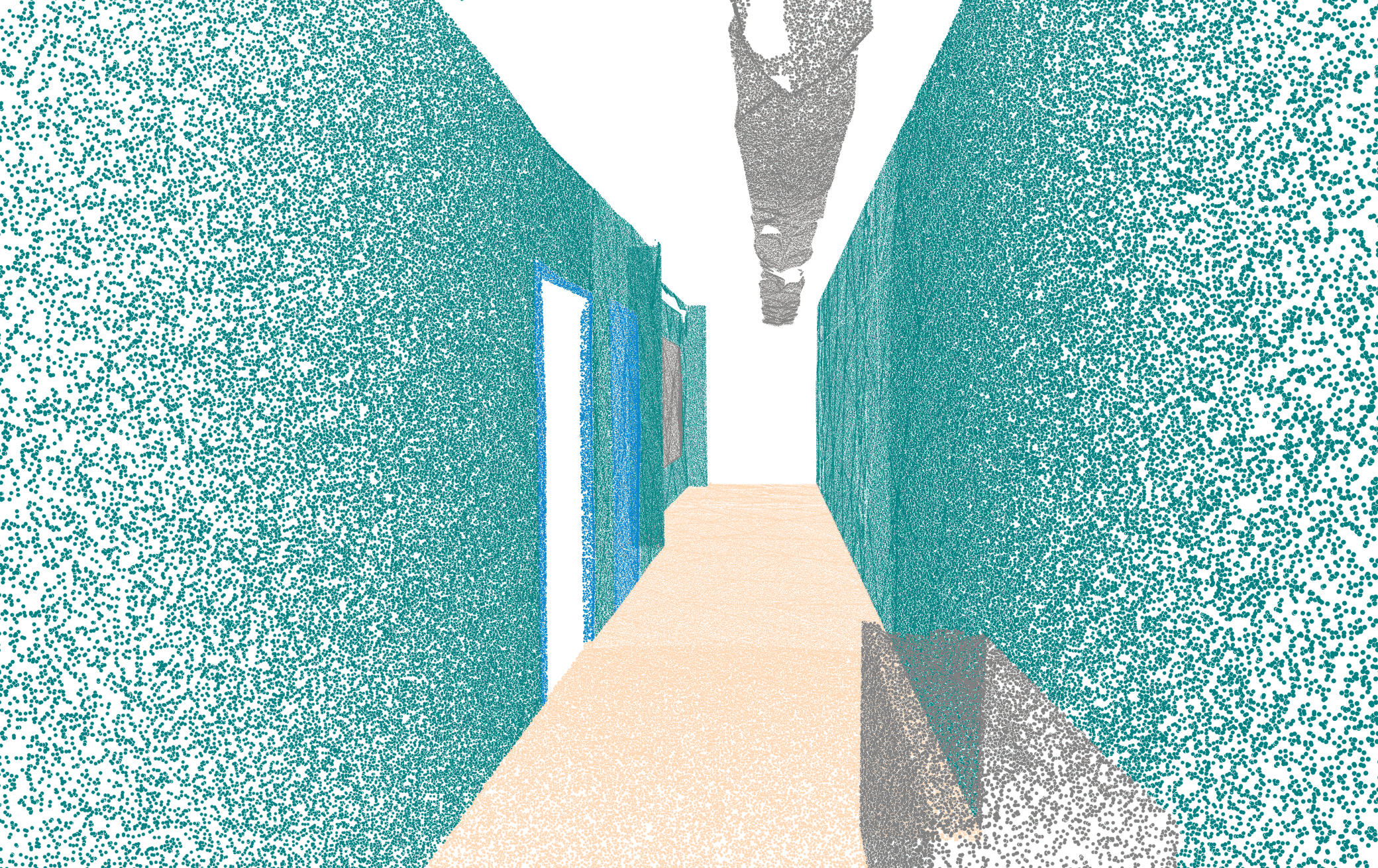}
    			\end{subfigure}
    			\begin{subfigure}{0.33\linewidth}
    				\centering
    				\includegraphics[width=1\linewidth]{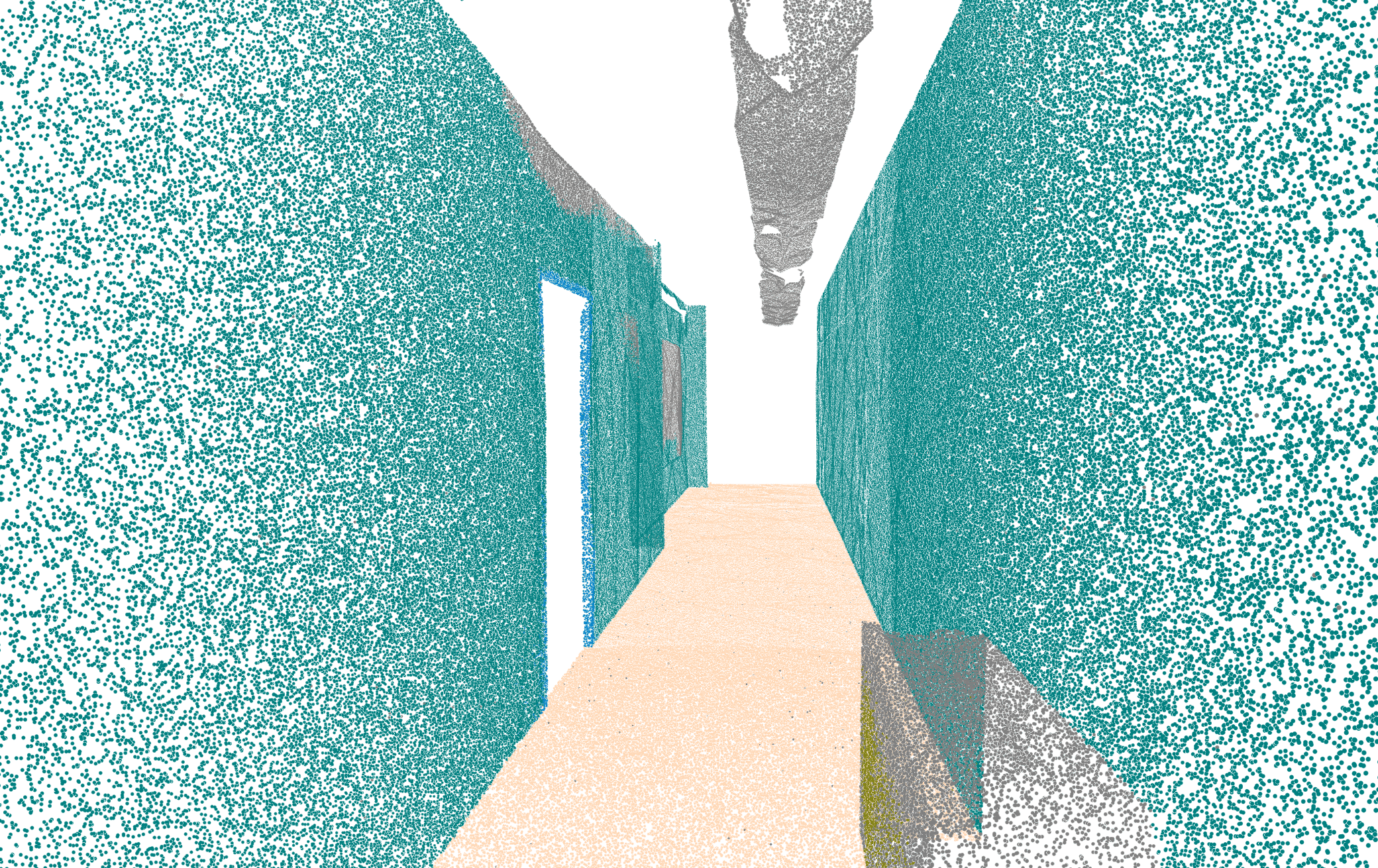}
    			\end{subfigure} \\
    			\begin{subfigure}{0.33\linewidth}
    				\centering
    				\includegraphics[width=1\linewidth]{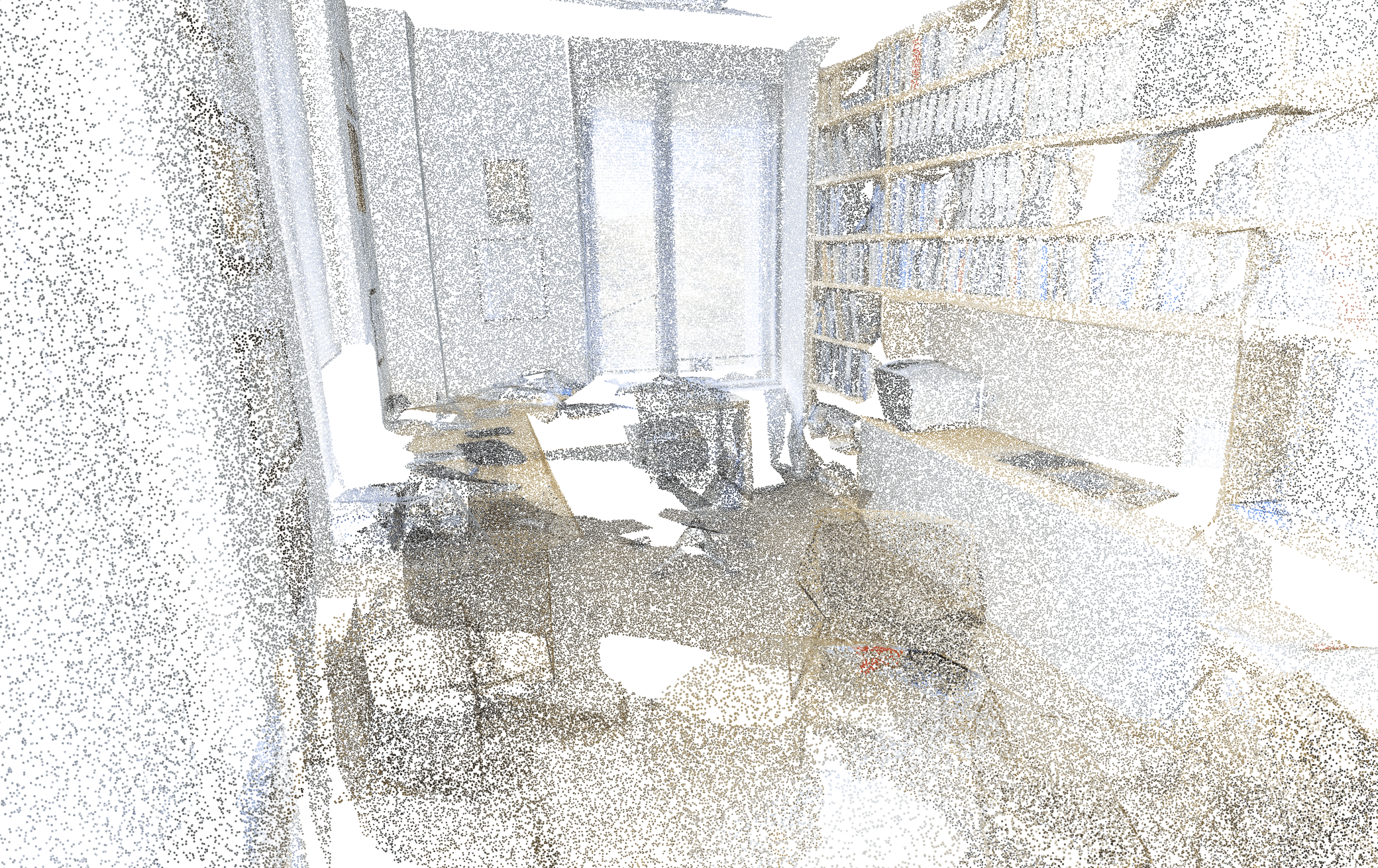}
    			\end{subfigure} \hfill
    			\begin{subfigure}{0.33\linewidth}
    				\centering
    				\includegraphics[width=1\linewidth]{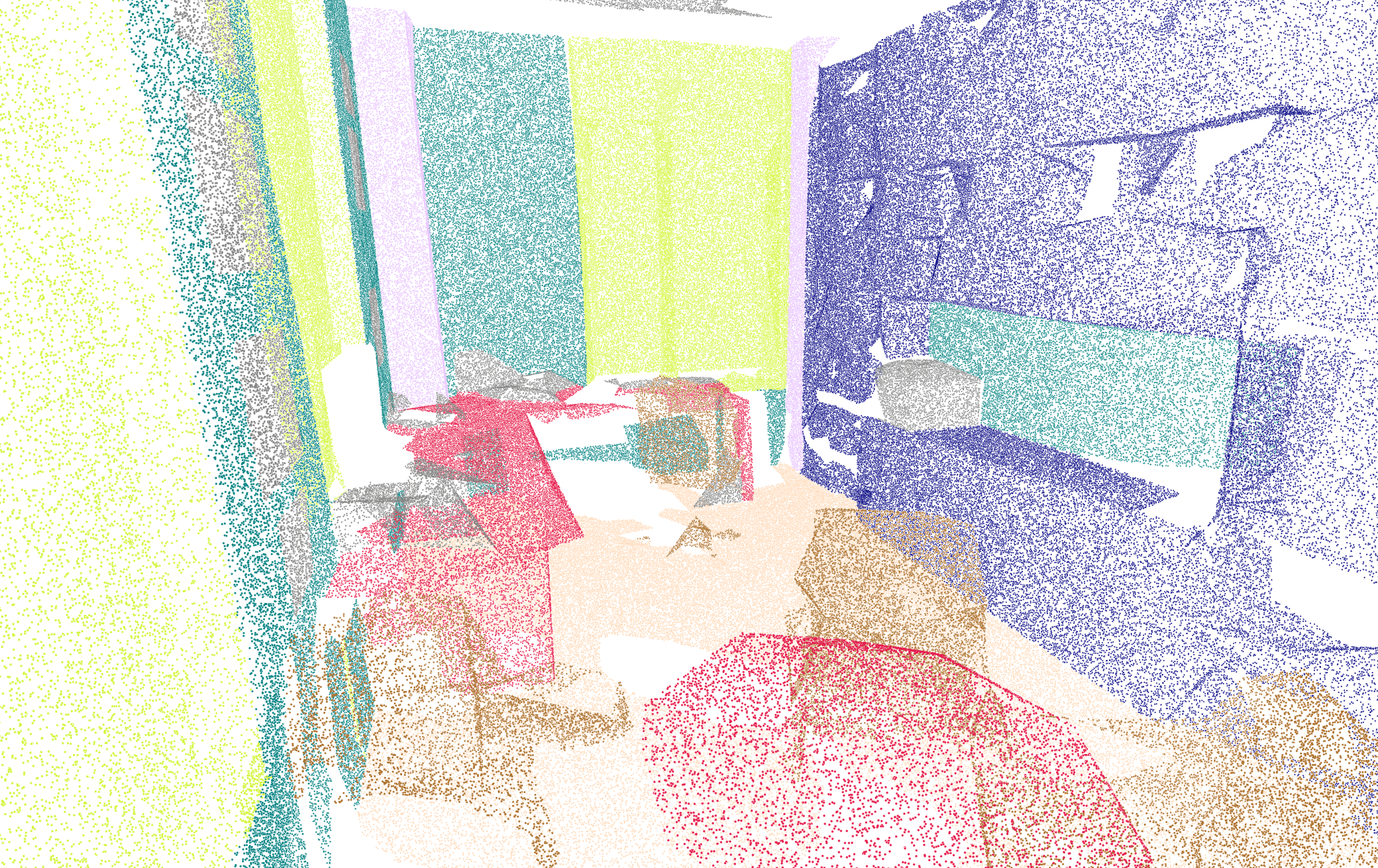}
    			\end{subfigure}
    			\begin{subfigure}{0.33\linewidth}
    				\centering
    				\includegraphics[width=1\linewidth]{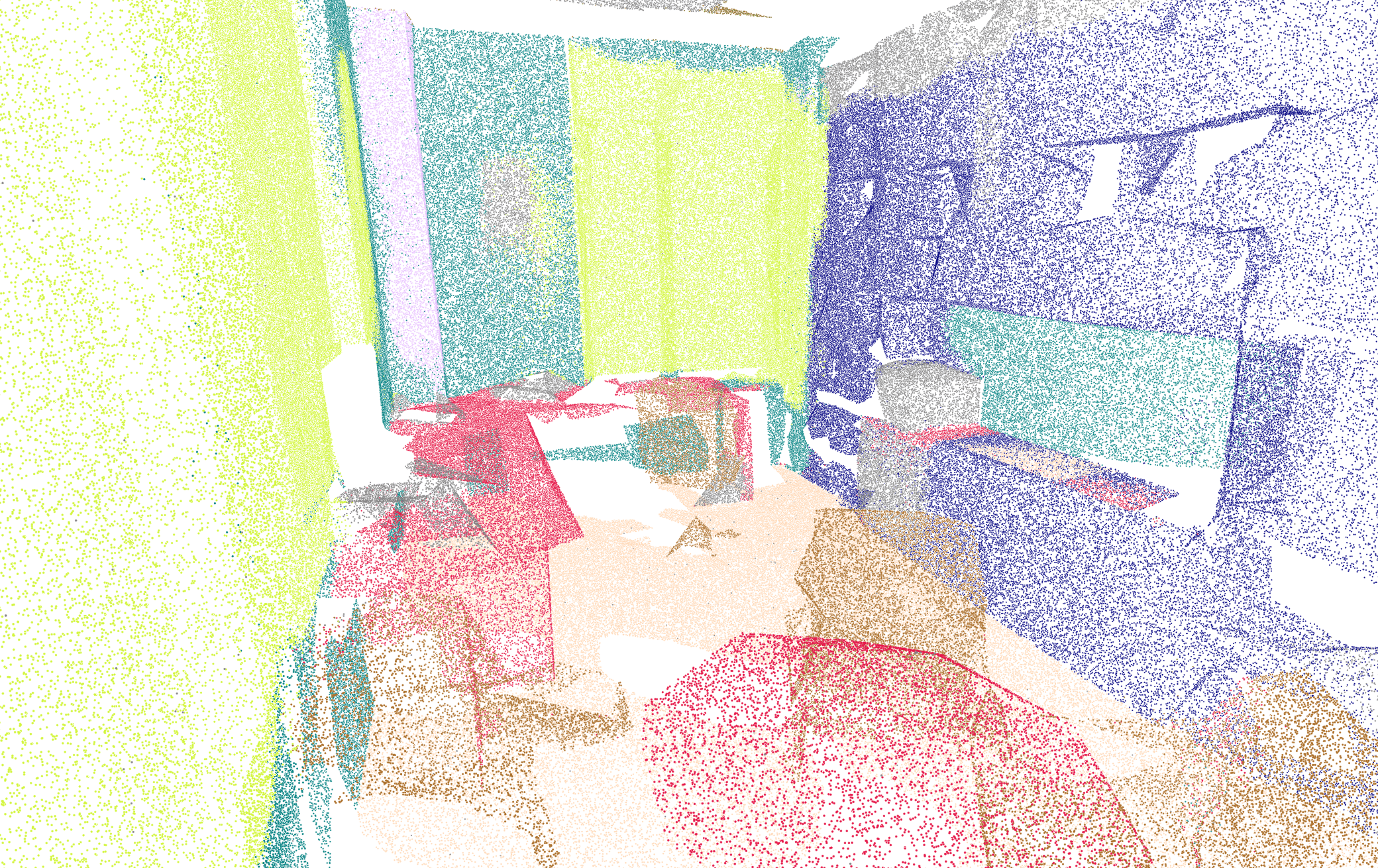}
    			\end{subfigure} \\
    			\begin{subfigure}{0.33\linewidth}
    				\centering
    				\includegraphics[width=1\linewidth]{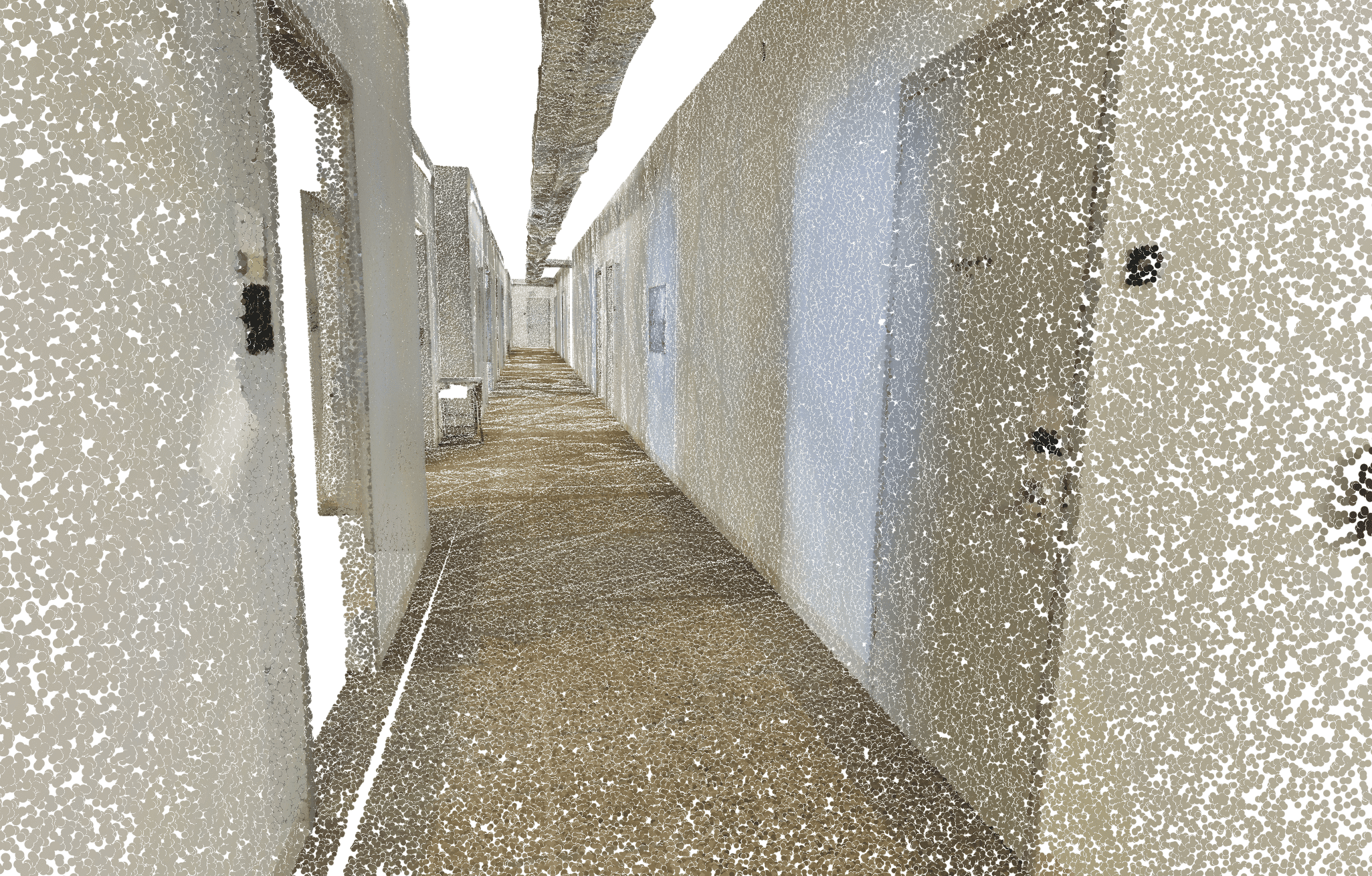}
    			\end{subfigure} \hfill
    			\begin{subfigure}{0.33\linewidth}
    				\centering
    				\includegraphics[width=1\linewidth]{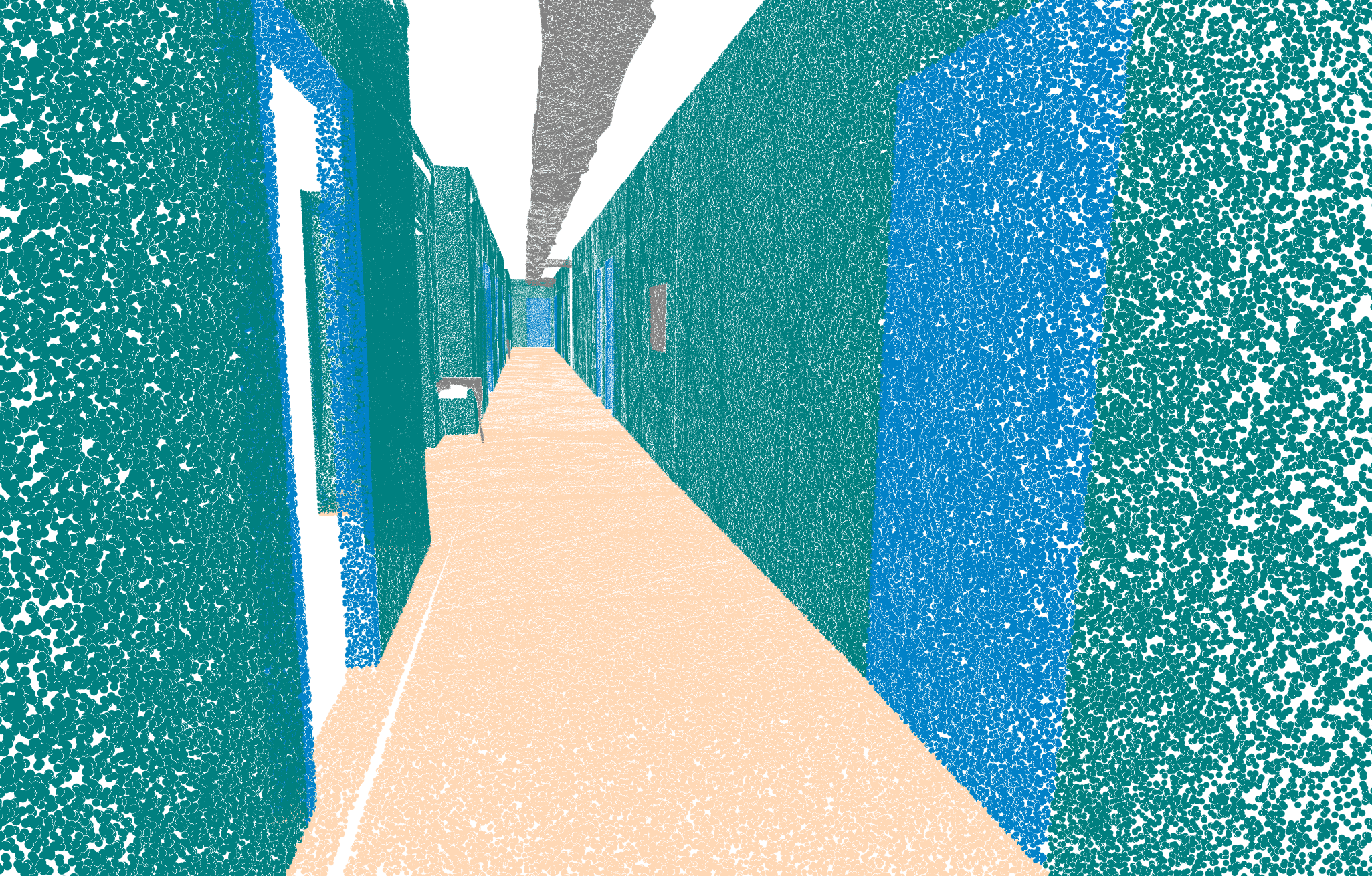}
    			\end{subfigure}
    			\begin{subfigure}{0.33\linewidth}
    				\centering
    				\includegraphics[width=1\linewidth]{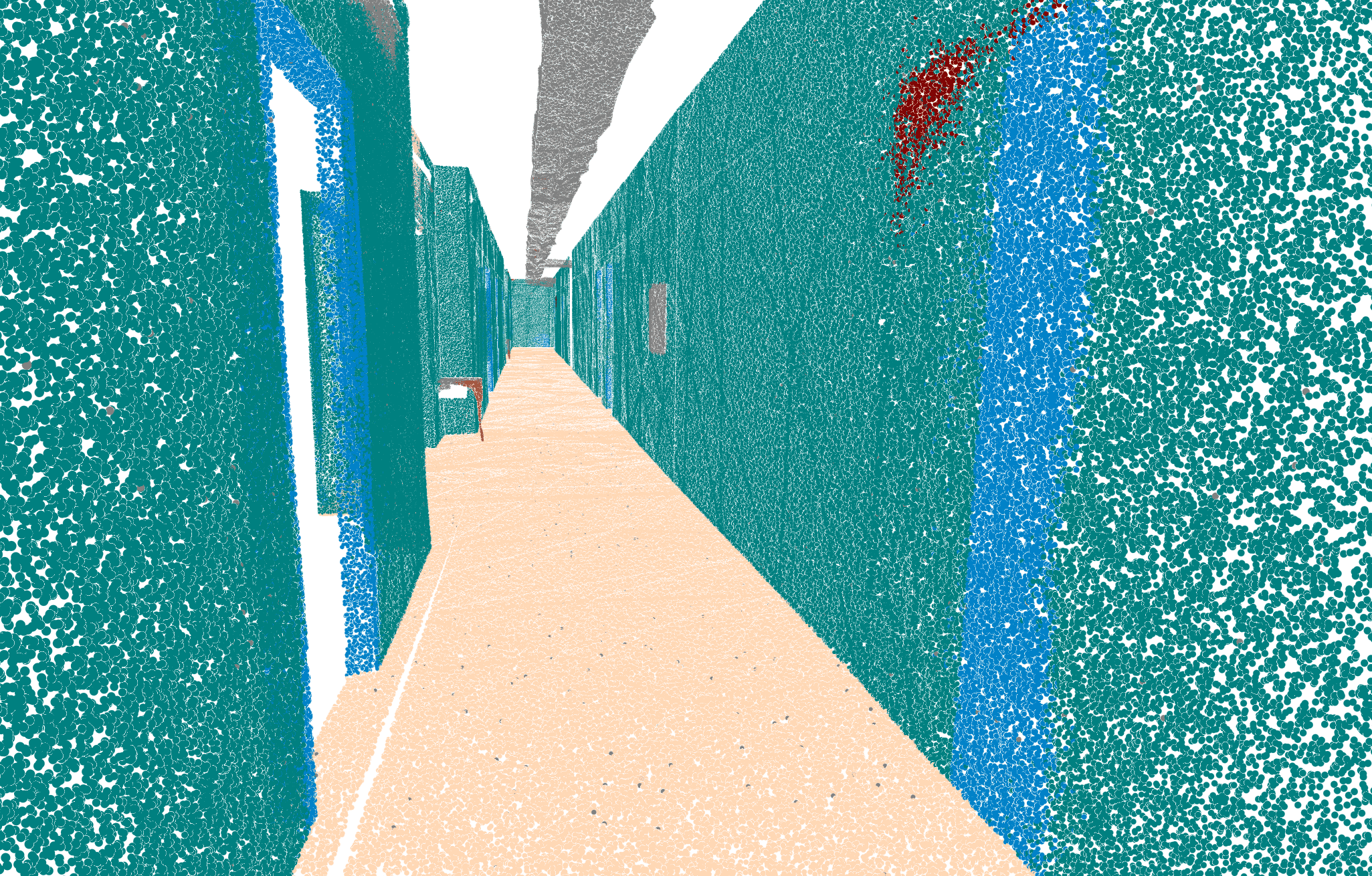}
    			\end{subfigure} \\
    			\begin{subfigure}{0.33\linewidth}
    				\centering
    				\includegraphics[width=1\linewidth]{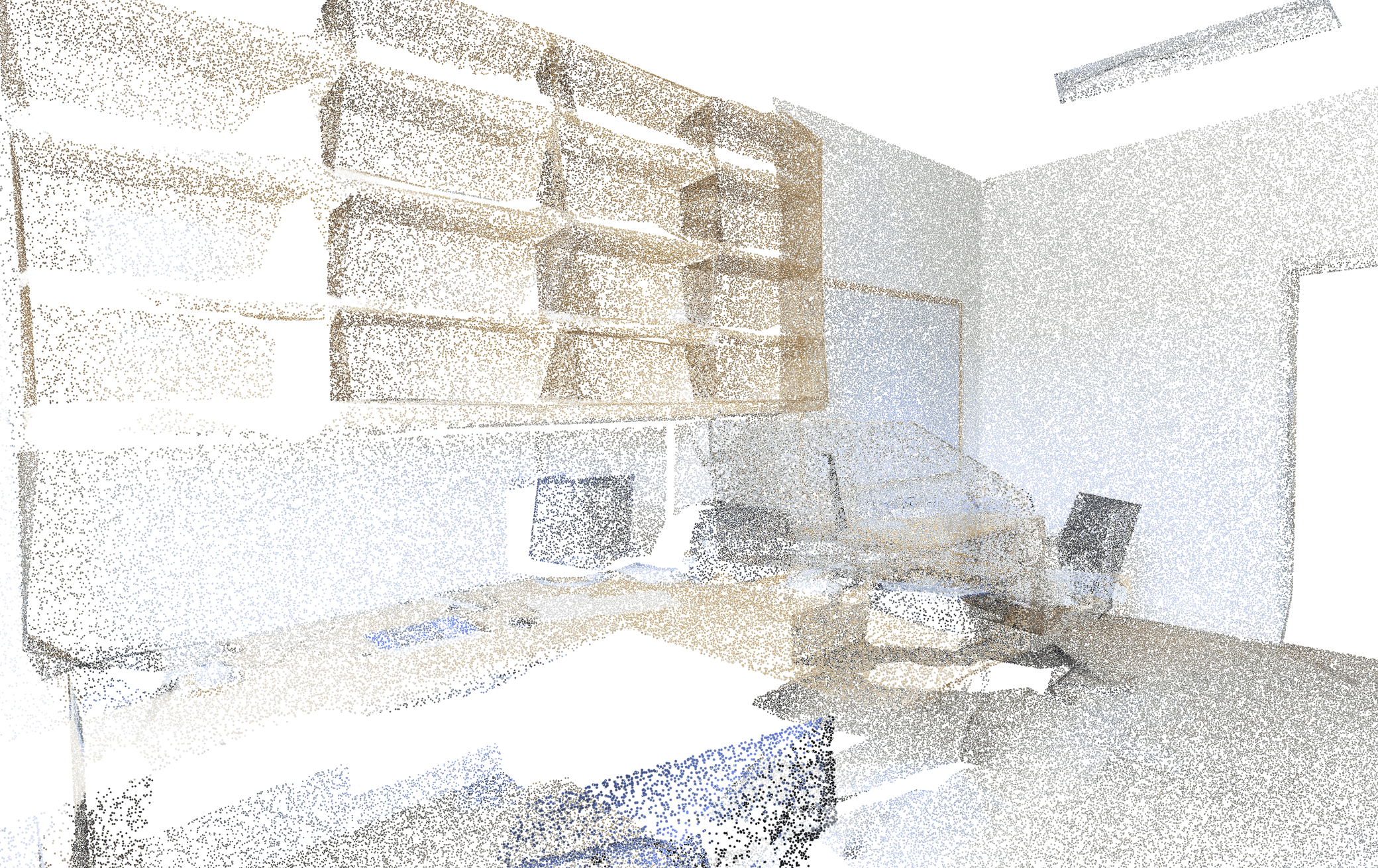}
    				\caption{Input XYZ+RGB}        
    			\end{subfigure} \hfill
    			\begin{subfigure}{0.33\linewidth}
    				\centering
    				\includegraphics[width=1\linewidth]{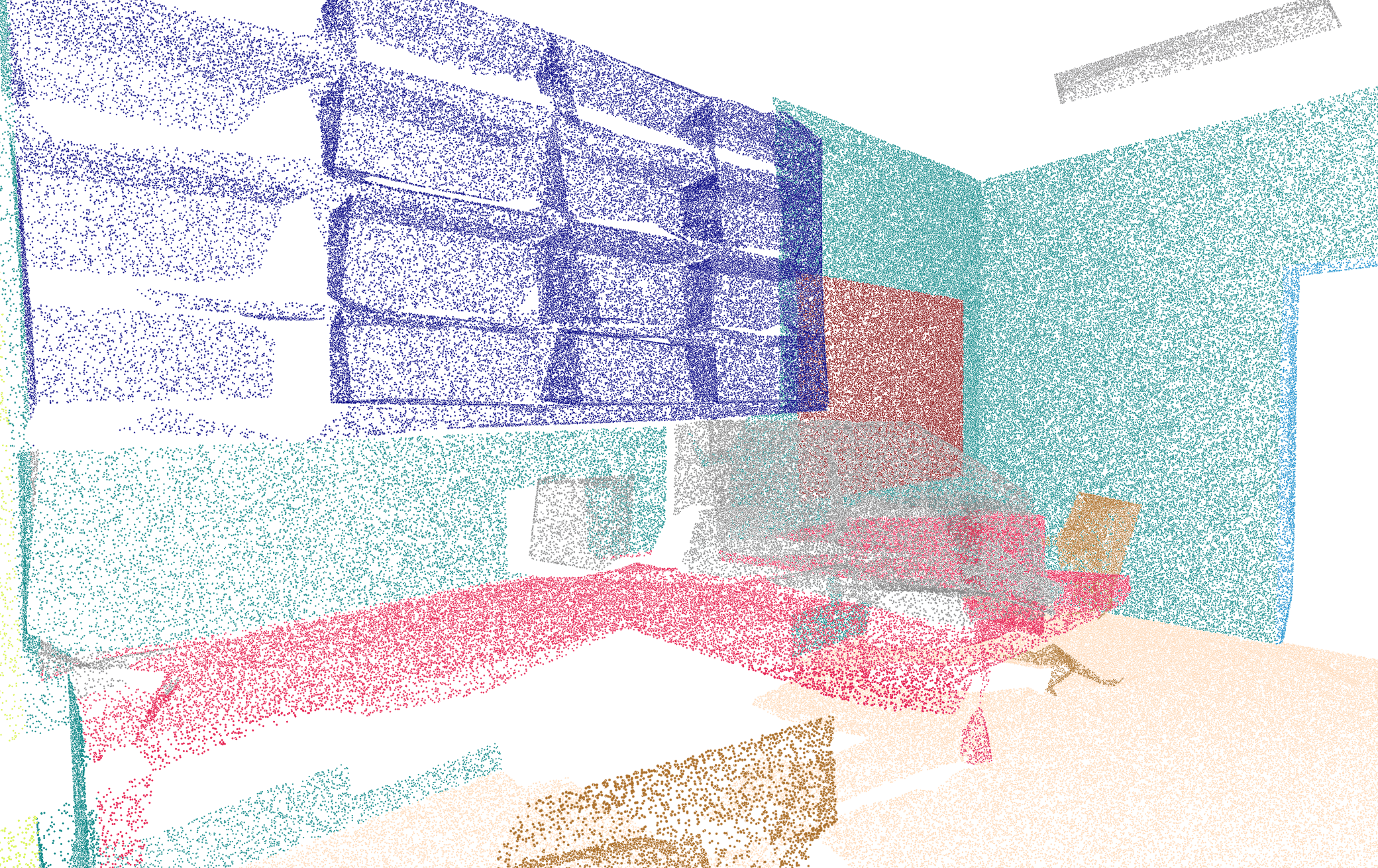}
    				\caption{Ground truth}
    			\end{subfigure}
    			\begin{subfigure}{0.33\linewidth}
    				\centering
    				\includegraphics[width=1\linewidth]{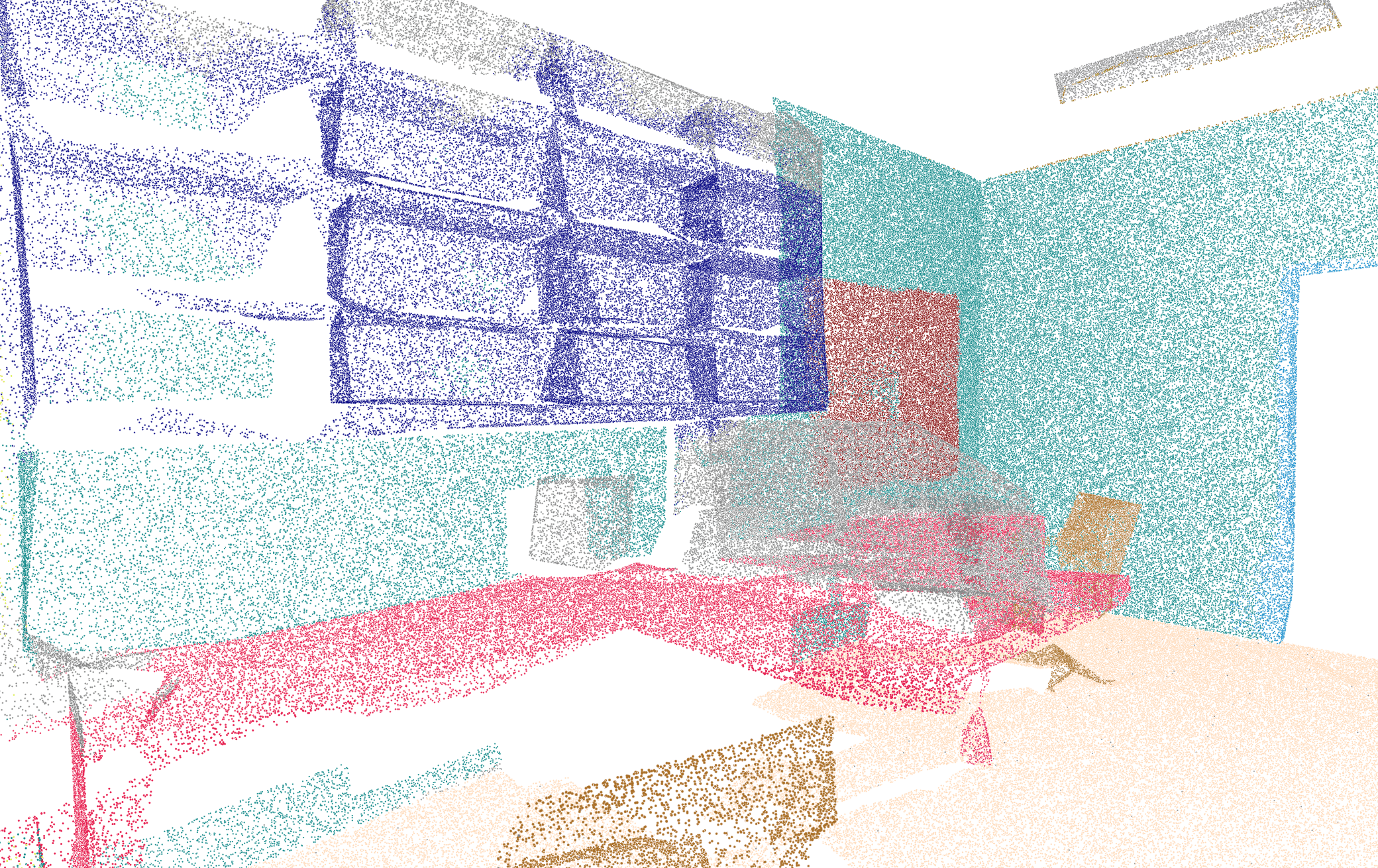}
    				\caption{Ours}
    			\end{subfigure} 
    			\captionsetup{aboveskip=5pt}
    			\captionsetup{belowskip=0pt}
    		\end{adjustwidth}
    		\caption{More visual result of S3DIS\cite{2017arXiv170201105A} area 5}
    		\label{fig:more2}
    	\end{figure*}

    	We report the IOU of each object class in Table \ref{tb:ciou} and visualize more results of S3DIS\cite{2017arXiv170201105A} in Figure \ref{fig:more1} and \ref{fig:more2}. The segmentation results are generated by our full model. In the visual results, Grid-GCN can predict objects such as chairs and tables very accurately, but sometimes mislabels the points on the border of two planar objects such as a board and a wall.
    	\vspace{-10pt}
    	\begin{table*}[h!]
    		\setlength\tabcolsep{2.5pt}
    		\begin{tabular}{lcc|ccccccccccccc}
    			\hline
    			Method            & OA             & mIoU           & ceiling        & floor          & wall           & beam          & column         & window         & door           & table          & chair          & sofa           & bookcase       & board          & clutter        \\ \hline
    			PointNet\cite{qi2017pointnet}          & -              & 41.09          & 88.80          & 97.33 & 69.80          & \textbf{0.05} & 3.92           & \textbf{46.26}          & 10.76          & 58.93          & 52.61          & 5.85           & 40.28          & 26.38          & 33.22          \\
    			SegCloud\cite{tchapmi2017segcloud}          &                & 48.92          & 90.06          & 96.05          & 69.86          & 0.00          & \textbf{18.37}          & 38.35          & 23.12          & 70.40          & 75.89          & \textbf{40.88}          & 58.42          & 12.96          & 41.60          \\
    			PointCNN\cite{li2018pointcnn}          & 85.91          & 57.26          & 92.31          & \textbf{98.24}          & \textbf{79.41} & 0.00          & 17.60          & 22.77          & \textbf{62.09} & \textbf{74.39}          & 80.59          & 31.67          & 66.67          & 62.05          & \textbf{56.74} \\
    			\textbf{Grid-GCN} & \textbf{86.94} & \textbf{57.75}          & \textbf{94.12} & 97.28          & 77.66          & 0.00          & 16.61          & 32.91          & 58.53          & 72.15          & \textbf{81.32} & 36.46          & \textbf{68.74} & \textbf{64.54} & 50.46          \\ \hline
    		\end{tabular}
    		\caption{Segmentation result on S3DIS\cite{2017arXiv170201105A} area 5. We report overall accuracy (OA, $\%$),  mean class IoU (mIoU, $\%$) and per-class IoU ($\%$). Grid-GCN achieves the highest overall accuracy and mIoU among 4 models.}
    		\label{tb:ciou}
    	\end{table*}
    	\vspace{-15pt}
    	\pagebreak
    	\section{Summary of notations, acronyms and concepts}
    	In the paper, we introduce the following concepts as well as their definitions. Here we list them again for clarity. 
    	\begin{table*}[htb!]
    		\setlength\extrarowheight{1pt}
    		\begin{tabular}{ll}
    			\hline
    			Notations and Concepts      & Descriptions       \\ \hline
    			\multirow{2}{*}{}GridConv & \multirow{2}{*}{}A network layer of our model, which includes a data structuring stage by using   \\ 
    			&  Coverage-Aware Grid Query and a data aggregation stage by using Grid Context Aggregation.    \\ \hline
    			CAGQ                &   A data structuring module, named Coverage-Aware Grid Query.             \\ \hline
    			GCA                &    A graph convolution module, named Grid Context Aggregation.             \\ \hline
    			RPS                &    Random    Point Sampling: A method that randomly samples $M$ group centers from $N$ points.         \\ \hline
    			\multirow{2}{*}{}FPS & \multirow{2}{*}{}Farthest Point Sampling: A method that samples one point a time,  \\ 
    			&  each time picks the point that maximizes    the distance to the selected points.    \\ \hline
    			RVS                &    Random Voxel Sampling (CAGQ)            \\ \hline
    			CAS                &    Coverage-Aware    Sampling (CAGQ)            \\ \hline
    			$N$                &  The number of input points of a GridConv layer.     \\ \hline
    			\multirow{2}{*}{}$M$ & \multirow{2}{*}{}The number of point groups sampled from $N$ input points, the Grid Context Aggregation module  \\  &   aggregates node points' information to each group center, then output $M$ representative points.    \\ \hline
    			$K$                &  The number of node points in each point groups.  \\ \hline
    			center voxel            &  For each point group, we select an occupied voxel and query node points in its neighborhood.   \\ \hline
    			$O_v$                &   The set of all occupied (non-empty) voxels in the space.  \\ \hline
    			$O_c$                &   The set of $M$ center voxels that's sampled from $O_v$.  \\ \hline
    			node points                &   For each point group, we query $K$ points from context points in a neighborhood.  \\ \hline
    			group center            &   The barycenter of $K$ node points in each point group.  \\ \hline
    			$\pi (v_i)$                &   The occupied voxel neighbors of an occupied voxel $v_i$.  \\ \hline
    			$\lambda$            &   The number of occupied voxel neighbors of an occupied voxel $v_i$.  \\ \hline
    			$n_v$                &   The max number of points CAGQ stores in each occupied voxel.          \\ \hline
    			\multirow{2}{*}{}context points & \multirow{2}{*}{}All stored points in $\pi (v_i)$, these points are the context points of the center voxel $v_i$  \\  &   and the point group sampled in this neighborhood afterwards.      \\ \hline
    			$\chi_i$                &   The x,y,z vector of a node point $p_i$ or a group center $c$.         \\ \hline
    			\multirow{2}{*}{}$w_i$ & \multirow{2}{*}{}The coverage weight of a node point $p_i$, which is the number of points that have been aggregated \\
    			&   to that point in previous layers. We initialize $w_i$ to 1 for each raw points.  \\ \hline
    			$f_i$                &   The semantic features carried by a node point $p_i$.         \\ \hline \\[-11pt]
    			$\widetilde{f}_i$                &   The semantic features calculated from $f_i$.         \\ \hline 
    			$e_i$                &   The edge attention between node point $p_i$ and the center, calculated by the edge attention function. \\ \hline \\[-11pt]
    			$\widetilde{f}_{c,i}$                &   The contribution from $\widetilde{f}_i$ to the center, determined by $e_i$ and $\widetilde{f}_i$.         \\ \hline \\[-11pt]
    			$\widetilde{f}_c$                &   The features of the group center, aggregated from all node points' contribution $\widetilde{f}_{c,i}$.  \\ \hline
    		\end{tabular}
    		\caption{Notations, acronyms and concepts.}
    	\end{table*}
    \end{appendices}

\end{document}